%% file: egpaper_for_review.tex
\documentclass[10pt,twocolumn,letterpaper]{article}

\usepackage{cvpr}
\usepackage{times}
\usepackage{epsfig}
\usepackage{graphicx}
\usepackage{amsmath}
\usepackage{amssymb}
\usepackage{UserCommands}
\usepackage{makecell}
\usepackage{dsfont}
\usepackage{enumitem}
\usepackage{subfigure}
\usepackage{caption}
\usepackage{arydshln}
\usepackage[pagebackref=true,breaklinks=true,letterpaper=true,colorlinks,bookmarks=false]{hyperref}

\cvprfinalcopy % *** Uncomment this line for the final submission

\def\cvprPaperID{8067} % *** Enter the CVPR Paper ID here

\ifcvprfinal\fi

\begin{document}

%%%%%%%%% TITLE
\title{D3Feat: Joint Learning of Dense Detection and Description of 3D Local Features}
% \title{Joint Learning of Dense Keypoint Detection and Description for 3D Point Clouds}
% JDDNet: Joint Detection and Description for 3D Local Features
% End-to-End Learning 3D Keypoint Detector and Descriptors.

\author{Xuyang Bai$^{1}$\hspace{0.5cm} Zixin Luo$^{1}$\hspace{0.5cm} Lei Zhou$^{1}$\hspace{0.5cm} Hongbo Fu$^{2}$\hspace{0.5cm} Long Quan$^{1}$\hspace{0.5cm}  Chiew-Lan Tai$^{1}$\hspace{0.5cm} \\
\normalsize $^1$Hong Kong University of Science and Technology \hspace{0.7cm} $^2$City University of Hong Kong \hspace{0.7cm} \normalsize  \\
\tt\small\{xbaiad,zluoag,lzhouai,quan,taicl\}@cse.ust.hk \hspace{0.7cm}
\tt\small hongbofu@cityu.edu.hk}

\maketitle
%\thispagestyle{empty}

%%%%%%%%% ABSTRACT
\begin{abstract}
% In this paper, we propose the JDDNet, a fully convolutional network for 3D point cloud which is simultaneously a keypoint detector and a feature descriptor.  Unlike most previous works which solely focus on keypoint description, we observed that detection and description play complementary roles in point cloud registration, indicating potential joint improvement. Therefore, we design a density-invariant selection strategy to detect keypoints on point cloud using dense feature map, thus integrate a detector and a descriptor into one single network. \xyc{"thus" logic unclear? maybe replace it by "and further integrate ..."} We also propose a novel descriptor-guided score loss to self-supervise the detector, which dramatically improve the detector performance. We perform extensive experiments on both indoor and outdoor datasets to validate our approach. By jointly optimization, the obtained descriptor out-performs the state-of-the-art methods by a significant margin when using small number of keypoints, and the detector also achieves state-of-the-art performance in terms of keypoint repeatability.
% \hbc{2018 template? }
A successful point cloud registration often lies on robust establishment of sparse matches through discriminative 3D local features. Despite the fast evolution of learning-based 3D feature descriptors, little attention has been drawn to the learning of 3D feature detectors, even less for a joint learning of the two tasks. In this paper, we leverage a 3D fully convolutional network for 3D point clouds, and propose a novel and practical learning mechanism that densely predicts both a detection score and a description feature for each 3D point. In particular, we propose a keypoint selection strategy that overcomes the inherent density variations of 3D point clouds, and further propose a self-supervised detector loss guided by the on-the-fly feature matching results during training. Finally, our method achieves state-of-the-art results in both indoor and outdoor scenarios, evaluated on 3DMatch and KITTI datasets, and shows its strong generalization ability on the ETH dataset. Towards practical use, we show that by adopting a reliable feature detector, sampling a smaller number of features {is} sufficient to achieve accurate and fast point cloud alignment. 
[\href{https://github.com/XuyangBai/D3Feat/}{code release}]
\end{abstract}

%%%%%%%%% BODY TEXT
\section{Introduction}
    \input{Introduction.tex}
%------------------------------------------------------------------------
\section{Related Work}
\input{RelatedWork.tex}
    
%-------------------------------------------------------------------------
\section{Joint Detection and Description Pipeline}
\label{section:methodology}
    \input{Methodology_1.tex}

%------------------------------------------------------------------------
\section{Joint Optimizating Detection \& Description}
    \input{Methodology_2.tex}

%------------------------------------------------------------------------  
\section{Implementation Details}
    \input{Implementation.tex}

%------------------------------------------------------------------------ 
\section{Experiments}
   \input{Experiment.tex}

%------------------------------------------------------------------------ 
\section{Conclusion}
    \input{Conclusion.tex}

%------------------------------------------------------------------------ 
\newpage
{\small
\bibliographystyle{ieee}
\bibliography{egbib}
}

\newpage 
\setcounter{page}{1}
\section{Supplementary Material}
  \input{Supplementary.tex}

\end{document}

%% file: Introduction.tex
Point cloud registration aims to find {an optimal} transformation between two partially overlapped point cloud fragments, which is a fundamental task in applications such as simultaneous localization and mapping (SLAM)~\cite{montemerlo2002fastslam, salas2013slam++} and 3D Lidar-based mapping~\cite{schonberger2016structure}. In above contexts, the local keypoint detection and description serve as two keys for obtaining robust point cloud alignment results.

\indent 
The recent research on 3D local {feature descriptors} has shifted to learning-based approaches. However, due to the difficulty of acquiring ground-truth data, most existing works often overlook the keypoint detection learning in point cloud matching, and instead randomly sample a set of points for feature description.
% \zyc{explain the reason of overlooking more clearly? The reason here is quite confusing for outliers} \xyc{I think it is not necessary as we will discuss some work on keypoint detection in next paragraph, so maybe no need to explain why it is overlooked.} 
Apparently, this strategy might suffer {from} several drawbacks. First, the {randomly sampled} points are {often poorly localized}, {resulting} in inaccurate transformation estimates during geometric verification such as RANSAC~\cite{fischler1981random}.
% \hbc{well most of the readers might know RANSAC well, it's still better to give a reference here}. 
Second, those random points might appear {in non-salient regions like smooth surfaces, which may lead to} 
%such as smooth surface, which resulting in 
indiscriminative descriptors that adversely introduce noise in later matching steps. Third, to obtain a full scene coverage, an oversampling of random points is required that considerably decreases the efficiency of the whole matching process. In essence, we argue that a small number of keypoints suffice to align point clouds successfully, 
%suffices to solve for geometry, 
and well-localized keypoints can 
%enable a more accurate geometry recovery.
further improve the registration accuracy. % In this spirit, a reliable keypoint detector is sought to ...
The imbalance of detector and descriptor learning motivates us to learn these two tightly coupled components jointly.\\
% a keypoint detector and a feature descriptor.}\\
% \zyc{Seems not that straightforward to get this conclusion here}
 %joint with a keypoint descriptor whose capabilities are tightly coupled.
% \noindent \taic{move one sentence from end of previous paragraph}
% \indent Learning a keypoint detector is non-trivial due to the lack of ground truth training data. 3DFeat-Net~\cite{jian20183dfeat} is a patch-based network which predicts an attention score and descriptor for one input point cloud patch, whereas only restricted spatial context can be used and the computational time proportional to the number of points is consumed during inference. Another attempt made by USIP~\cite{li2019usip} utilizes unsupervised training to encourage the keypoints to be invariant under transformation. They focus solely on keypoint detection, thus hampers joint improvement for combined learning of keypoint detector and descriptor. \xy{And both of their network are unable to densely predict the detection scores either, which severely limits the usage in applications that requires high resolutional point cloud input.} Hence, to mitigate the limitations of existing works, in this paper, we propose a novel method to learn a keypoint detector and descriptor in a joint manner. To the best of our knowledge, we are the first to handle densely feature detection and description jointly on 3D point cloud.\\
\indent However, the learning-based 3D keypoint detector has not received much attention in previous studies. One attempt made by 3DFeat-Net~\cite{jian20183dfeat} predicts a patch-wise detection score, whereas only limited spatial context is considered and a dense inference for the entire point cloud is not applicable in practice. Another recent work USIP~\cite{li2019usip} adopts an unsupervised learning scheme that encourages {keypoints} to be covariant under arbitrary transformations. However, without a joint learning of detection and description, the resulting {detector might} not match the capability of {the} descriptor, thus preventing the release of the potential for a fully learned 3D feature. Instead, in this paper, we seek for a joint learning framework that is able to not only predict keypoints densely, but also tightly couple {the detector} with a descriptor with shared weights for fast inference. \\
\indent To this end, we draw inspiration from D2-Net~\cite{dusmanu2019d2} in 2D domain for a joint  learning of a feature detector and descriptor. However, the extension of D2-Net for 3D point clouds is non-trivial. First, a network that allows for dense feature prediction 
% \hbc{of what?} 
in 3D is needed instead of previous patch-based architectures.
% \hbc{Why this is challenging for D2-Net?}\xyc{Dense feature extraction is not challenging for 2D images(include D2-Net), but is not common for 3D point cloud. Previous work usually consume a point cloud patch, and predict the feature for the central point.}.
In this work, we resort to KPConv~\cite{thomas2019kpconv}, {a newly proposed convolutional operation on 3D point clouds,} to build a fully convolutional network to consume {an unstructured 3D point cloud directly}.
% \hbc{cite 3DFeat-Net?}\xyc{But most of the previous works on 3D descriptors consume point patches as the input, should we only cite 3DFeat-Net?}
Second, we adapt D2-Net to handle the inherent density variations of 3D point clouds, which is the key to achieve highly repeatable keypoints in 3D domain. Third, observing that the original loss in D2-Net does not guarantee convergence in our context, we propose a novel self-supervised detector loss guided by the on-the-fly feature matching results during training, so as to encourage the detection scores to be consistent with the reliability of predicted keypoints.  To summarize, our contributions are threefold: 

\begin{enumerate}[itemsep=-1mm]
\vspace{-0.1cm}
    \item[1.] We leverage a fully convolutional network based on KPConv, and adopt a joint learning framework for 3D local feature detection and description, without constructing dual structures, for fast inference.
    % We design a 3D fully convolutional network that densely predicts a detection score and description feature for each 3D point.
    % \item[1.] We extend a keypoint selection strategy to train a fully convolutional network that has a dual role of both keypoint detector and dense feature descriptor for 3D point clouds.   
    \item[2.] We propose a novel density-invariant keypoint selection strategy, which 
    % to evaluate the keypoint detection score \tai{"score" is used twice} of points 
    is the key to obtaining repeatable keypoints for 3D point clouds. %due to the inherent density variations.}
    %, which overcomes the inherent density variations of 3D point cloud.
    \item[3.] We propose a self-supervised detector loss that receives meaningful guidance from the on-the-fly feature matching results during training, which guarantees the convergence of tightly coupled descriptor and detector.
    % \item[4.] \tai{this is result} Our model can perform dense feature extraction rapidly with state-of-the-art discriminative power. We achieve state-of-the-art performance on 3DMatch and KITTI benchmark using much {fewer} keypoints.
\end{enumerate}

We demonstrate the superiority of our method {over the state-of-the-art methods} by conducting extensive experiments on both 3DMatch of indoor settings, and KITTI, ETH of outdoor settings. {To our best knowledge, we are the first to handle the \textbf{D}ense \textbf{D}etection and \textbf{D}escription of 3D local \textbf{Feat}ures for 3D point clouds in a joint learning framework. We refer to our approach as \Name.}
%To our best knowledge, we are the first to handle the {dense} feature detection and description for 3D point clouds in a joint learning framework. 
%\xy{We refer to our approach for \textbf{D}ense \textbf{D}etection and \textbf{D}escription of 3D local \textbf{Feat}ures as \Name.}
% \hbc{In the abstract you mention `several' benchmarks. `Several' should be `two'?}

%% file: RelatedWork.tex
\subsection{3D Local Descriptors}
\vspace{-0.1cm}
Early approaches to extract 3D local descriptors are mainly hand-crafted \cite{Guo2016}, which generally lack robustness against noise and occlusion. To address this, recent studies on 3D descriptors have shifted to learning-based approaches, which is the main focus of our paper.
% Unlike 2D regular grids, the sparse and unstructured nature of point clouds makes traditional neural networks not applicable.  
% Several {3D data representations} have been proposed for learning local geometric features in 3D data.

\noindent \textbf{Patch-based networks.} Most existing learned local descriptors require point cloud patches as input. Several 3D data representations have been proposed for learning local geometric features in 3D data. 
Early attempts like~\cite{su2015multi, zhou2018learning} use multi-view image representation for descriptor learning. Zeng et al.~\cite{zeng20173dmatch} and Gojcic et al.~\cite{gojcic2019perfect} convert 3D patches into a voxel grid of  truncated distance function~(TDF) values and smoothed density value~(SDV) representation respectively. Deng et al.~\cite{deng2018ppfnet, deng2018ppf} build their network upon PointNet to directly consume unordered point sets. 
% Early attempts like~\cite{su2015multi, zhou2018learning} first project 3D patches into {the image domain} and then use the well-studied 2D convolutional neural networks (CNNs) to extract feature descriptors. Zeng et al.~\cite{zeng20173dmatch} convert 3D patches into a {truncated distance function} (TDF) representation and use 3D convolution to extract feature descriptors. 3DSmoothNet~\cite{gojcic2019perfect} uses a voxelized smoothed density value (SDV) representation to learn compact descriptors and achieves the state-of-the-art performance. PointNet~\cite{qi2017pointnet} is a seminal work that directly consumes a raw unstructured point cloud as input. Built upon PointNet, {the method by} Deng et al.~\cite{deng2018ppfnet, deng2018ppf} uses point pair features to learn a rotation-invariant local descriptor for point cloud patches. 
% \hbc{Use 1-2 sentences to discuss the limitations of the above approach or how they are related to our work.}
Such patch-based methods suffer from efficiency problem as the intermediate network activations are not reused across adjacent patches, thus severely limits their usage in applications that requires high resolutional output.

\noindent \textbf{Fully-convolutional networks.} Although fully convolutional networks introduced by Long et al.~\cite{long2015fully} have been widely used in the 2D image domain, it has not been extensively explored in the context of 3D local descriptor. {Fully convolutional geometric feature (FCGF)}~\cite{Choy_2019_ICCV} is the first to adopt a fully convolutional setting for dense feature description on point clouds. It uses sparse convolution proposed in~\cite{choy20194d} to extract feature descriptors and achieves rotation invariance by simple data augmentation.  However, their method does not handle keypoint detection. 
%\xyc{where should this paragraph be?} \tai{keep it here. Discussion is a bit brief and not clear why most similar. Should elaborate. }

%-------------------------------------------------------------------------
\subsection{3D Keypoint Detector}
\vspace{-0.1cm}

Unlike the exploration of learning-based 3D local descriptors, most existing methods for 3D keypoint detection are hand-crafted. A comprehensive review of such methods can be found in~\cite{tombari2013performance}. The common trait among hand-crafted approaches is their reliance on local geometric properties of 
%\tai{the points for keypoints selection (unclear. what are "the points"?)}. \xyc{should be "rely on local geometric properties of 
point clouds. Therefore, severe performance degradation occurs when such detectors are applied to real-world 3D scan data where noise and occlusion commonly exist. 
% 3DFeatNet~\cite{jian20183dfeat} is the first learning-based network that learns feature detector and descriptor using weak supervision. However, the attention layer that predicts the salience of each point is the by-product of the learning descriptors, thus the performance of keypoint detection is not guaranteed.\xyc{I will explain 3Dfeat in the next paragraph in detail, should I delete the intro here?} \tai{yes, it should be removed from here. No repetition. Be concise.}
% In contrast, we will propose a detector network capable of detecting keypoints with desired repeatability and distinctiveness for matching. 
To make the detector learnable, {Unsupervised Stable Interest Point (USIP)}~\cite{li2019usip} proposes an unsupervised method to learn keypoint detection. {However, USIP is unable to densely predict the detection scores and its network has the risk to be degenerate if the desired keypoint number is small. {In contrast,} our network is able to predict dense detection scores without having the risk of degeneracy.}
% Different from their work, we observed the potential of joint improvement for combined learning of keypoint detector and descriptor.

%-------------------------------------------------------------------------
\subsection{Joint Learned Descriptor and Detector}
\vspace{-0.1cm}

In 2D image matching, {several works have tackled} the keypoint detection and description problems jointly~\cite{yi2016lift, detone2018superpoint, luo2018geodesc, ono2018lf, christiansen2019unsuperpoint, revaud2019r2d2, luo2019contextdesc}. However, adapting these methods to the 3D domain is challenging and less explored.
%have not been extensively explored. 
As we know, 3DFeat-Net~\cite{jian20183dfeat} is the only work that attempts to jointly learn the keypoint detector and descriptor for 3D point clouds. However, their method focuses more on learning the feature descriptor with an attention layer estimating the salience of each point patches as its by-product, {and} thus the performance of {their} keypoint detector is not guaranteed. Besides, their method takes point patches as input, which is inefficient as addressed before.
% Besides, during inference, {their method} needs to construct {a} cluster centered at each point to get the attention score and feature vector for each point, \xy{which is not efficient as the computation of adjacent patches do not share.}
% Thus {the time complexity of their approach is} 
%it requires computational time 
% proportional to the number of the involved points.
In contrast, we aim to detect keypoint locations and extract per-point features in a single forward pass for efficient usage.
%we will present an efficient 
% \tai{remove "time", the word "efficient" already means efficient in terms of time} 
%model that 
% can {efficiently} detect keypoint locations and extract per-point features in one forward pass 
% \hbc{If it is per-point, the computational time is still proportional to the number of points? Is there any comparison to show our method is more efficient?}\xyc{Our method predicts a detection score and a feature for each point in the point cloud by \textbf{one forward pass} for normal size point cloud(no matter how many points are there), but 3DFeatNet(all patch-based methods) consumes a local patch near one point to predict the score and feature for this single point, so if there are n point in the point cloud, it has to do forward pass n times to get the scores of all the points (although you can input multiple patches one time, it still proportional to the number of points due to memory limitation. Maybe I should clarify this difference in Sec 2.2 }
% . In addition, our proposed network serves a dual role by fusing 
% {a} detector network and {a} descriptor network into a single network, thus saving memory and computation consumption by a large margin.
Specifically, our proposed network serves a dual role by fusing 
the detector and descriptor network into a single one, thus saving memory and computation consumption by a large margin.
% \hbc{Again better to have a comparison to support such a claim}. \xyc{I think it is clear as we have a single network for both detection and description so the memory and computation are shared, but other works have two separate networks. }

%% file: Methodology_1.tex
Inspired by {D2-Net}, a recent approach {proposed} by Dusmanu et al. for 2D image matching~\cite{dusmanu2019d2}, instead of training {separate networks for keypoint detection and description,}
%a separate keypoint detection network, 
we design a single neural network that plays a dual role: a dense feature descriptor and a feature detector. However, adapting {the idea of D2-Net}
%this idea 
to the 3D domain is non-trivial because of the irregular {nature} and varying sparsity of point clouds. In the following, we will first describe the fundamental steps to perform feature description and detection on irregular 3D point clouds, {and} then explain our strategy of dealing with the sparsity variance
% \hbc{how about the irregular issue?}\xyc{The irregular issue are tackled by KPConv as it directly consume the the raw point cloud as input, maybe I shouldn't mention the irregular nature of 3D point clouds since KPConv is not our novelty?}
in {the} 3D domain. 
\begin{figure*}[ht]
	\vspace{-0.5cm}
    % \begin{minipage}[t]{0.5\linewidth}
        % \centeringf
        \includegraphics[width=8.5cm]{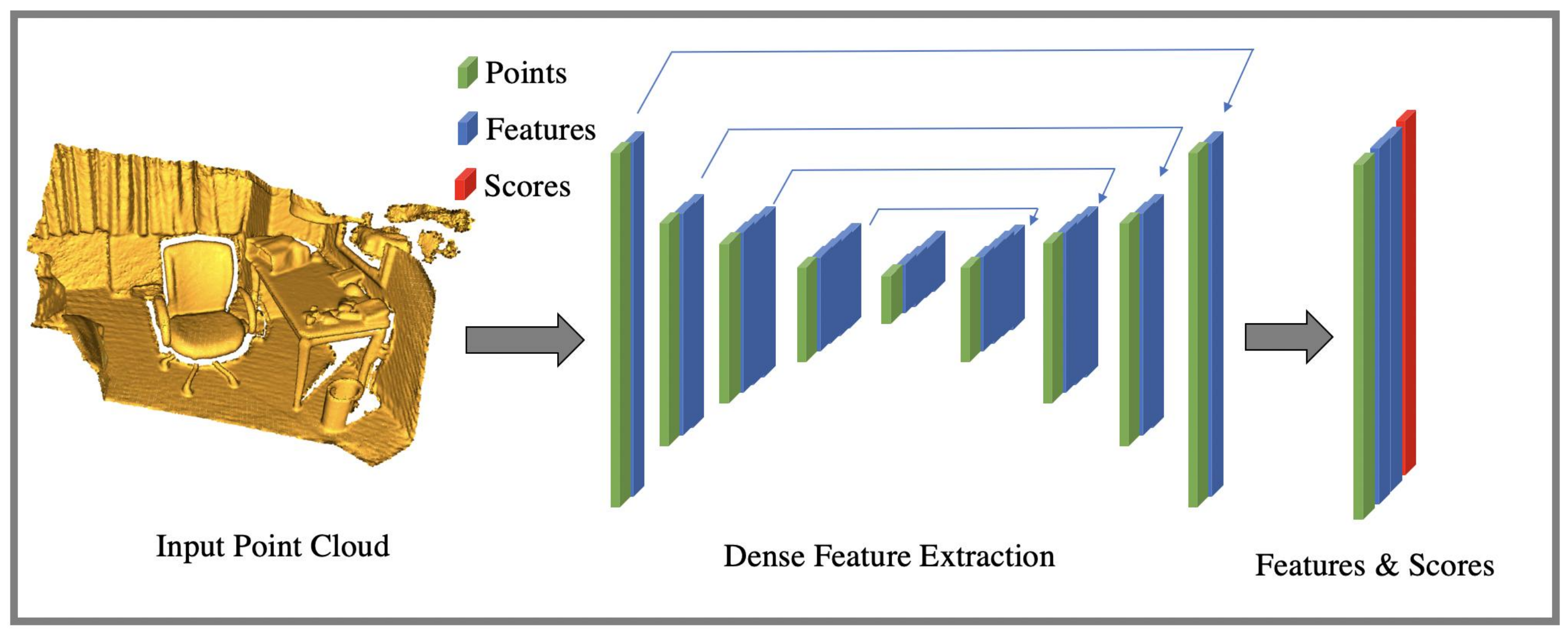}
        % \label{fig:pipeline:a}
    % \end{minipage}
    % \begin{minipage}[t]{0.5\linewidth}
        % \centering
        \includegraphics[width=8.85cm]{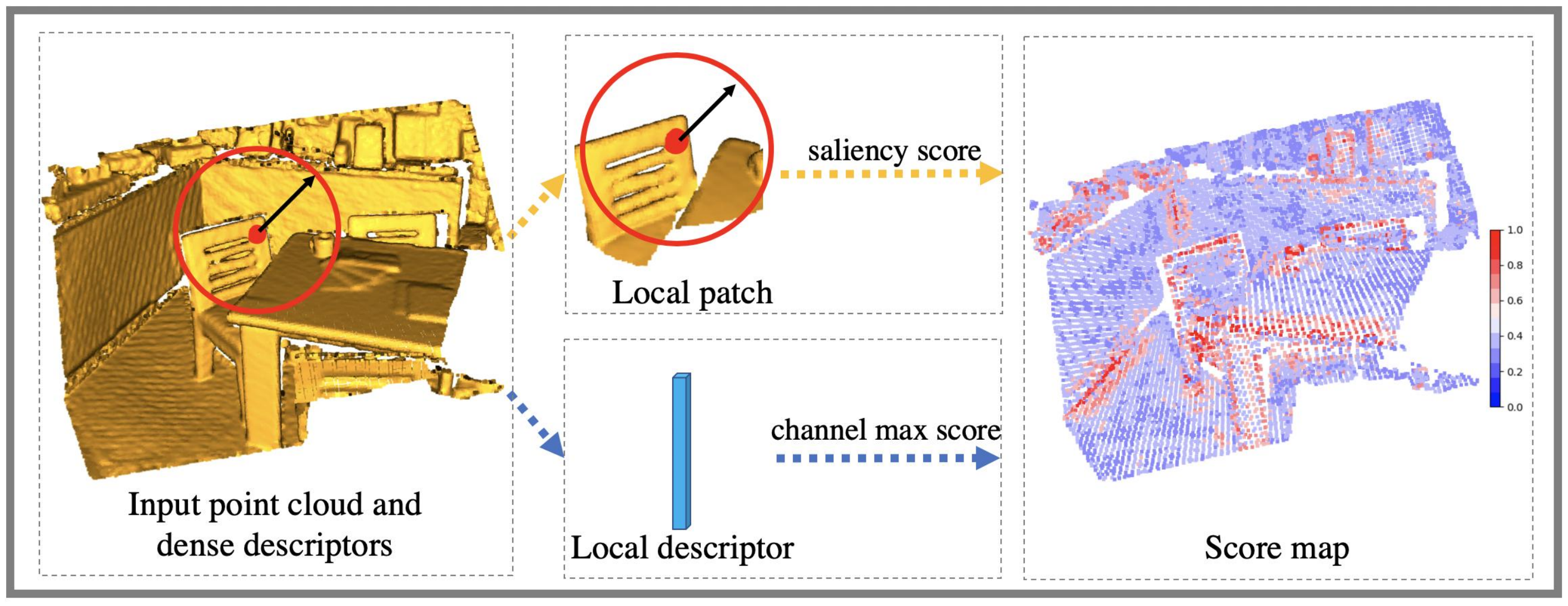}
        % \label{fig:pipeline:b}
    % \end{minipage}
    \caption{({Left}) The network architecture of \Name. Each block indicates a ResNet block using KPConv to replace image convolution. All layers except the last one are followed by batch normalization and ReLU. ({Right}) Keypoint detection. After dense feature extraction, we calculate the keypoint detection scores by applying saliency score and channel max score. This figure is best viewed with color and zoom-in. } 
    \label{fig:pipeline}
\end{figure*}

\subsection{Dense Feature Description}
To address the issue of convolution on irregular point clouds and better capture the local geometry information, Thomas et al. proposed the Kernel Point Convolution~(KPConv)~\cite{thomas2019kpconv}, which uses kernel points carrying convolution weights to emulate the kernel pixels in 2D convolution, {and then defines} the convolution operation on the raw point clouds. 
% We adapt KPConv and use it as our backbone network to perform dense feature extraction. 
{We adpot KPConv as our backbone network to perform dense feature extraction.}
Below we first briefly review the formulas of KPConv.
% The network architecture is illustrated in Fig~\ref{fig:pipeline}. \xyc{ maybe I should delete this sentence, it seems more suitable to refer to this fig in the following paragraph} \tai{yes, understanding the previous sentence doesn't need a figure. better to move the reference to figure to where the details of the architecture are explained.}

Given a set of points $P \in \mathbb{R}^{N \times 3}$ and a set of features $F_{in} \in \mathbb{R}^{N \times D_{in}}$ represented in a matrix form, let $x_i$ and $f_i$ denote {the $i$-th point in $P$ and its corresponding feature in $F_{in}$, respectively}. The general convolution by kernel $g$ at point $x$ is defined as
    \begin{equation}
        (F_{in} * g) = \sum_{x_i \in N_{x}} g(x_i - x)f_i,
        % \taic{equations should be part of sentences, need a comma here. Please check throughout}
        \label{eqn1}
    \end{equation}
    where $N_{x}$ is the radius neighborhood of point $x$, and $x_i$ is a supporting point in this neighborhood. The kernel function is defined as
    \begin{equation}
        g(x_i-x) = \sum_{k=1}^Kh(x_i-x,\hat x_k)W_k,
        \label{eqn2}
    \end{equation}
    where $h$ is the correlation function between the kernel point $\hat x_k$ and the supporting point $x_i$, $W_k$ is the weight matrix of the kernel point $\hat x_k$, and $K$ is the number of kernel points. 
    We refer readers to the original paper~\cite{thomas2019kpconv} for more details.  \\
    % defined as truncated linear correlation function:
    % \begin{equation}
    %     h(y_i,\hat x_k) = max(0, 1 - \frac{||y_i - \hat x_k||}{\sigma})
    % \end{equation}
    % where $\sigma$ is the influence distances of kernel points. 
    \indent The original formulation of KPConv is not invariant to point density. 
    % The authors claimed that, by using grid subsampling strategy, they can ensure the spatial consistency of point sampling locations and improve the robustness to varying densities. While {this} is true for dense indoor scenes, point clouds from outdoor Lidar scans still suffer from density variation with regular subsampling strategy. 
    Thus we add a density normalization term, {which sums up the number of supporting points in the neighborhood of $x$}, to Equation~\ref{eqn1} to ensure that convolution is sparsity invariant:
    \begin{equation}
        (F_{in} * g) = \frac{1}{|N_x|}\sum_{x_i \in N_{x}} g(x_i - x)f_i.
         \label{eqn3}
    \end{equation}
    % From our experiment, we found it extremely useful to outdoor dataset where Lidar scans have huge sparsity changes. \xy{I remove }\\
    \indent Based on the normalized kernel point convolution, we adopt a UNet-like structure with skip connections and residual blocks to build a fully convolutional network~\cite{long2015fully}, {as illustrated in 
    %. Our network architecture is illustrated in 
    Fig.~\ref{fig:pipeline} (Left)}.  
    % \xyc{I add this sentence to provide some connection between KPConv and our fully convolutional network} \tai{too subtle, intended connection to KPConv not obvious}
    
    % Unlike~\cite{deng2018ppf, deng2018ppfnet}, where local patches near each keypoint {are} extracted and processed by a neural network to generate the descriptor for each keypoint,
    Unlike patch-based methods which only support sparse feature description,
    % \hbc{the contrast is not clear. You mean the previous methods only support sparse feature description? Why?}\xyc{Yes, previous patch-based method is not applicable for dense description because of time complexity.}, 
        our network is able to
    %\taic{no claims until you present evidence. When describing methodology, focus on explaining design rationale} 
    perform {dense} feature description under {a} fully convolutional setting. The output of our network is a dense feature map in the form of a two-dimensional matrix $F \in \mathbb{R}^{N \times c}$, where $c$ is the dimension of the feature vector. The descriptor associated with point $x_i$ is denoted as $d_i$,
    \begin{equation}
        d_i = F_{i:},~ d_i \in \mathbb{R}^c,
        % \taic{full stop here. Check throughout}
         \label{eqn4}
    \end{equation}
    where $F_{i:}$ denotes the $i$-th row of two-dimensional matrix $F$.
    % \hbc{Previously you said $F$ is a dense feature vector}.\xyc{$F$ is a dense feature map in form of a 2-dimensional matrix (I am not sure which word is suitable, in 2D they usually say feature map or feature vector)}
    The descriptors are L2-normalized to unit length.

\subsection{Dense Keypoint Detection}
    % In~\cite{dusmanu2019d2} \taic{Author et al. is better, gives readers an idea which paper, Author et al detect (active voice more concise and easier to read)}, keypoints on 2D images are detected based on the local maximum across the spatial dimensions of the feature maps and  the channel dimension, 
    Dusmanu et al.~\cite{dusmanu2019d2} detect keypoints on 2D images based on the local maximum across  the spatial and channel dimensions of the feature maps, and use a softmax operator to evaluate the \textit{local-max} score of a pixel. {Due to the regular structure of images, their approach simply selects the neighboring pixels as the neighborhood. To extend their approach to 3D, this strategy might be replaced by radius neighborhood to handle the non-uniform sampling setting of point clouds.} However,
%    {adapting} this idea to 3D domain is nontrivial \taic{this sentence has been mentioned too many times.} 
    %2D images are regular so the neighborhood of a pixel is well-defined, and each pixel have the same size of neighbourhood \tai{too trivial, remove or make more concise}. 
 %   {2D images are regular grids, so each pixel has well-defined neighborhood of the same size.}
    %In contrast, in 3D we usually use radius neighborhood instead of knn neighborhood since knn is not robust in non-uniform sampling setting. Nevertheless, 
    the number of neighboring points in the radius neighborhood can vary greatly. 
    % For indoor scenes where point cloud scans are dense, this problem can be partially handled by uniform subsampling strategies, but the points near the boundary of the scene still have less \tai{fewer} points in their neighborhood. For outdoor scenes, this varying density problem is more common as the Lidar scans have large sparsity variation even under uniform subsampling.\taic{The long explanation up to here in this paragraph seems repeating the same reason already explained concisely in previous paragraph. Well-known fact should be stated very concisely, so that novelty in this paper can be quickly read}\xyc{may change another word to describe sparsity variation / varying density} 
    % \xy{I remove the explanation here, it should be easy to understand by saying "the boundary of indoor scenes, regions far away from Lidar center of outdoor scenes"}
{In this case,} if we simply use a  softmax function to evaluate the local maximum in the spatial dimension, the local regions with few points (e.g., {regions close to} the boundaries of indoor scenes or far away from Lidar center of outdoor scenes) would inherently have higher scores. To handle this problem, we propose a density-invariant saliency score to evaluate the saliency of a certain point compared with its local neighborhood. \\
    \indent Given the dense feature map $F \in \mathbb{R}^{N \times c}$, we regard $F$ as a collection of 3D response $D^k~(k=1,...,c)$:
    \begin{equation}
        D^k = F_{:k}, \ \ D^k \in \mathbb{R}^N,
        \label{eqn5}
    \end{equation}
    {where $F_{:k}$ denotes the $k$-th column of the two-dimensional matrix $F$.}
    % \hbc{Again vector or matrix? Be consistent}.
    %, ":k" means we pick the k column, I follow the notation of D2Net}
    % In this representation, these $c$ different 3D response $D^k$ are analogous to the DoG response maps in SIFT. By analoging the SIFT, we are going to find the local maximum as our keypoints. 
    The criterion of point $x_i$ to be a keypoint is
    % \tai{no comma here. in a sentence you wont put a common here, right?}
    \begin{equation}
        \begin{aligned}
        x_i\mathrm{~is~a~keypoint} \Longleftrightarrow & \ k = \arg\max_{t} \ D_i^t  \ \ \mathrm{and} \\
        & i = \arg\max_{j \in N_{x_i}} D_j^k, 
        % & D^k_i~is~a~local~max~in~D^k,
        \label{eqn6}
        \end{aligned}
    \end{equation}
    where $N_{x_i}$ is the radius neighborhood of $x_i$. This means that the most preeminent channel is first selected, and then verified by whether or not it is a maximum of its spatial local neighborhood on that particular response map $D^k$. During training, we soften the above process to make it trainable by applying two scores, {as illustrated in
    % The approach is illustrated in
    Fig.~\ref{fig:pipeline} (Right). The details are given below}.
    
    % \begin{figure}[t]
    % \centering
    % \includegraphics[width=8cm]{figures/detection.png}
    % \caption{After dense feature extraction, we calculate the keypoint detection scores by applying local peakedness score and channel max score. Best view with color and zoom-in.}
    % \label{fig:detection}
    % \end{figure}
    
    \noindent \textbf{{Density-invariant} saliency score.} This score aims to evaluate how salient a point is compared with other points in its local neighborhood. % And this score shares the similar idea with non-maximum-suppression in 2d keypoint detection methods as the most salient point will suppress the local max score of other points in the same local neighbor. 
    In D2-Net~\cite{dusmanu2019d2}, the score evaluating the \textit{local-max} is defined as
    \begin{equation}
        \alpha_{i}^k = \frac{\exp(D_i^k)}{\sum_{x_j \in N_{x_i}} \exp(D_j^k)}.
        \label{eqn7}
    \end{equation}
    %where $N_{x_i}$ is the radius neighborhood of $x_i$. 
    This formulation, however, is not invariant to sparsity. Sparse regions inherently have higher scores than dense areas because the scores are normalized by sum. We therefore design a density-invariant saliency score as follows:
    \begin{equation}
        \alpha_i^k = \ln(1 + \exp(D_i^k - \frac{1}{|N_{x_i}|}\sum_{x_j\in N_{x_i}}D_j^k)).
        \label{eqn8}
    \end{equation}
    In this formulation, the saliency score of a point is calculated as the difference between its feature and the mean feature of its local neighborhood. Thus it measures the relative saliency of a center point with respect to that of the supporting points in the local region. Besides, using the average response in place of sum (\textit{c.f.}, Equation~\ref{eqn7}) prevents the score from being affected by the number of the points in the neighborhood. {In our experiments (Section~\ref{section:Exp_keypoint})}, we will show that our {saliency score} significantly improves the network's ability to handle point cloud keypoint detection with varying density.
    
    \noindent \textbf{Channel max score.} This score is designed to pick up the most preeminent channel for each point:
    \begin{equation}
        \beta_{i}^k = \frac{D_i^k}{\max_{t}(D_i^t)}.
        \label{eqn9}
    \end{equation}
    Finally, both of the scores are taken into account for the final keypoint detection score:
    \begin{equation}
        s_i = \max_k(\alpha_i^k\beta_i^k).
        \label{eqn10}
    \end{equation}
    Thus after we obtain the keypoint score map
    % \hbc{can you show more visualization results of this map together with the selected keypoints?}\xyc{OK, but the score map in 3D is not as intuitive as that in 2D images(like the Figure 1 right), I will add more visualization in Supplementary Material. The visualization of keypoints are already shown in Experiment and also Supplementary Material.}
    of an input point cloud, we select points with top scores as keypoints.
    % points with top scores can be selected as the keypoints
    % Different with D2Net, during training we do not perform point-cloud level normalization because that will convert the true saliency score of points into the relative score. \xyc{add reason why remove this normalization.}
    % Since we are going to use the score as the weight during the training, such kind of normalization will harm the effect of scores.

%% file: Methodology_2.tex
Designing a proper supervision signal is the key to joint learning of a descriptor and a detector. In this section, We will first describe the metric learning loss for the descriptor, {and} then design a detector loss from a self-supervised perspective.
% guided by the on-the-fly feature matching results during training.

\noindent \textbf{Descriptor loss.} To optimize the descriptor network, many works attempt to use metric learning strategies, like contrastive loss and triplet loss. We will utilize the contrastive loss, since  from our experiments we have found it to give better convergence performance. As for the sampling strategy, we adopt the \textit{hardest in batch} strategy proposed in~\cite{mishchuk2017working} to make the network focus on hard pairs.  

\indent Given a pair of partially overlapped point cloud fragments $P$ and $Q$, and a set of $n$ {pairs of corresponding 3D points}. Suppose $(A_i, B_i)$ is a correspondence pair and the two points have their corresponding descriptors $d_{A_i}$ and $d_{B_i}$ and scores $s_{A_i}$ and $s_{B_i}$. The distance between a positive pair is defined as the Euclidean distance between their descriptors as follows:
    \begin{equation}
        d_{pos}(i) = ||d_{A_i} - d_{B_i}||_2.
        \label{eqn11}
    \end{equation}
    {The} distance between a negative pair is,
    \begin{equation}
        \begin{aligned}
        & d_{neg}(i) = \min\{||d_{A_i} - d_{B_j}||_2\}~ \mathrm{s.t.} ||B_j-B_i||_2 > R,
        \label{eqn12}
        \end{aligned}
    \end{equation}
    where $R$ is the safe radius, and $B_j$ is the hardest negative sample that lies outside the safe radius of the true correspondences. The contrastive margin loss {is} defined as 
    \begin{equation}
        \begin{aligned}
        L_{desc} & =  \frac{1}{n} \sum_{i} \big[ \max(0, d_{pos}(i) - M_{pos}) \\
         & + \max(0, M_{neg} - d_{neg}(i)) \big],
         \label{eqn13}
        \end{aligned}
    \end{equation}
    where $M_{pos}$ is the margin for {positive pairs} and $M_{neg}$ is the margin for {negative pairs}.
    
\noindent \textbf{Detector loss.} To optimize the detector network, we seek for a loss formulation that encourages the easily matchable correspondences to have higher keypoint detection scores than the correspondences which are hard to match. In~\cite{dusmanu2019d2}, Dusmanu et al. proposed an extension to the triplet margin loss to jointly optimize the descriptor and the detector:
    \begin{equation}
         L_{det} = \sum_{i} \frac{s_{A_i}s_{B_i}}{\sum_i s_{A_i}s_{B_i}} max(0, M + d_{pos}(i)^2 - d_{neg}(i)^2),
         \label{eqn14}
     \end{equation}
where $M$ is the triplet margin. They claimed that in order to minimize the loss, the detector network should predict high scores for most discriminative correspondences and vice versa. However, their loss does not provide explicit guidance for the score term, and experimentally we found that their origin loss formulation does not guarantee convergence in our context.
% \xy{Moreover, in practice the margin is usually set to be 1, which is rarely reached during training. Thus the term $(M + d_{pos}(i) - d_{neg}(i))$ is positive for most triplets, leading a negative gradient to the detection score of all the points.} 
% Experimentally, we also observed that the training process using their loss formulation is unstable. 
% Moreover, in practice the margin is usually set to be 1, which is rarely reached during training, thus will exclude most correspondences. For example, if all the correspondence satisfy $ -M < d_{pos}(i) - d_{neg}(i) < 0$, which is the common cases in the later stage of the training, the loss will be zero and never provide any guidance for the detector.

 \indent Thus we design a loss term to explicitly guide the gradient of the scores. From a self-supervised perspective, we use the on-the-fly feature matching results to evaluate the discriminativeness of each correspondence, which will guide the gradient flow of {the score of each keypoint}. If a correspondence is already matchable under the current descriptor network, we want the score to be higher and vice versa.
 %\xyc{"vice versa" means if a correspondence cannot be matched, we want the score to be lower. Does this usage not make sense?}. 
 Specifically, we define the detector loss as
    \begin{equation}
        L_{det} = \frac{1}{n}\sum_{i}\big[(d_{pos}(i) - d_{neg}(i))(s_{A_i} + s_{B_i}) \big].
        \label{eqn15}
    \end{equation}
Intuitively, if $d_{pos}(i)-d_{neg}(i)<0$, it indicates that this correspondence can be correctly matched using the nearest neighbor search, and the loss term will encourage the scores {$s_{A_i}$ and $s_{B_i}$}
%($s_{A_i}$, $s_{B_i}$) 
of the two points in the correspondence pair to be higher. 
% The lower $d_{pos}(i) - d_{neg}(i)$ is, the larger the gradient is.
Conversely, if $d_{pos}(i) - d_{neg}(i) > 0$, the correspondence is not discriminative enough for the current descriptor network to establish a correspondence, {and} thus the loss will encourage the scores $s_{A_i}$ and $s_{B_i}$ to be lower. In order to minimize the loss, the detector should predict higher scores for matchable correspondences and lower scores for non-matchable correspondences. 
% The previously mentioned situation where loss defined in D2net ceases to optimize will never occur using our loss formulation.

%% file: Implementation.tex
% \textbf{Network Architecture.}
\noindent \textbf{Training.} Our network is implemented in TensorFlow. During training, we use all the point cloud fragment pairs with more than 30\% overlap. For each pair of fragments $P$ and $Q$,  we establish the correspondence set by first randomly sampl{ing} $n$ anchor points from $P$, then apply{ing} the ground-truth transformation
% \xyc{why plural? there is only one transformation for each pair of fragments}
to these points, and perform{ing} nearest neighbor search in fragment $Q$. The correspondence is accepted only if the {Euclidean} distance between two points is less than a threshold. We use grid subsampling to control the number of points and ensure the spatial consistency of point clouds. 

For the first layer, we use $0.03m$ as our grid size. The neighborhood radius is 2.5 times the grid size of the current layer. The local neighborhood for keypoint detection is the same as the first radius neighborhood. For each point cloud fragment, we apply data augmentation including adding Gaussian noise with standard deviation 0.005, random scaling $\in [0.9, 1.1]$, and random rotation angle $\in [0^{\circ}, 360^{\circ})$ around an arbitrary axis. We use a batch size of 1 and correspondence number $n = 64$. 
% and the loss back-propagate only for fragment pairs with more than 32 correspondences. 
We use positive margin {$M_{pos} = 0.1$}, negative margin {$M_{neg} = 1.4$} for the contrastive loss. {The loss terms for the detector and the descriptor are equally weighted.}
% \hbc{how about the triplet margin $M$?}\xyc{I doesn't use triplet loss in my work, for the ablation study I trained the model using triplet loss with M = 1, should I report here?}
We optimize the network using the Momentum optimizer with an initial learning rate of 0.1, which is exponentially decayed every epoch, and the momentum is set to 0.98. The network converges in about 100 epochs. 

\noindent \textbf{Testing.} During testing, our keypoint detection module adopts the hard selection strategy formulated by Equation~\ref{eqn6} instead of soft selection to get better quality keypoints. This strategy emulates non-maximum suppression and avoids having the selected keypoints lie too close to each other.

% In addition, during training we normalize the feature vector by minus the point cloud wise max before calculate the keypoint detection score to avoid overflow.

% \noindent \textbf{State-of-the-art} For descriptor part, we adopt P

%% file: Experiment.tex
The following sections are organized as follows. First, we demonstrate our method~(\Name) regarding point cloud registration on 3DMatch dataset~\cite{zeng20173dmatch} (indoor settings) and KITTI~\cite{Geiger2013IJRR} dataset (outdoor settings), with the provided data split to train \Name. Next, we study the generalization ability of {\Name} on ETH dataset~\cite{Pomerleau:2012} (outdoor settings), using the model trained on 3DMatch dataset. Finally, we more specifically evaluate the keypoint repeatability on 3DMatch and KITTI datasets, to clearly demonstrate the performance of our detector.

\subsection{Indoor Settings: 3DMatch dataset}
We follow the same protocols in 3DMatch dataset~\cite{zeng20173dmatch} to prepare the training and testing data. In particular, the test set contains $8$  scenes with partially overlapped point cloud fragments and their corresponding transformation matrices. For each fragment, $5000$ randomly sampled 3D points are given for methods that do not include a detector.

\smallskip\noindent\textbf{Evaluation metrics.} Following~\cite{Choy_2019_ICCV}, two evaluation metrics are used including 1) \emph{Feature Matching Recall}, a.k.a. the percentage of successful alignment whose inlier ratio is above some threshold (i.e., $\tau_2=5\%$), which measures the matching quality during pairwise registration. 2) \emph{Registration Recall}, a.k.a. the percentage of successful alignment whose transformation error is below some threshold (i.e., RMSE $< 0.2$m), which better reflects the final performance in practice use. Besides, the intermediate result, \emph{Inlier Ratio}, is also reported. In above metrics 1) and 3), a match is considered as an inlier if the distance between its corresponding points is smaller than $\tau_1=10$cm under ground-truth transformation. For metric 2), we use a RANSAC with 50,000 max iterations~(following~\cite{zeng20173dmatch}) to estimate the transformation matrics.

\smallskip\noindent\textbf{Comparisons with the state-of-the-arts.}
We compare {\Name} with the state-of-the-arts on 3DMatch dataset. As shown in Table~\ref{tab:feature_matching_recall}, the \emph{Feature Matching Recall} is reported regarding two levels of data difficulty: the original point cloud fragments (left columns), and the randomly rotated fragments (right columns). In terms of \Name, we report the results both on $5000$ randomly sampled points (\NameRand) and on $5000$ keypoints predicted by our detector network (\NamePred). In both settings, {\Name} achieves the best performance when a learned detector is equipped.
% \tai{would be better to introduce a name for our proposed method so that future papers can refer to easily. Use def to define the name so that all occurrences of the name can be changed at once if needed} \xyc{The senior students said it will be better to introduce a name for our method after the paper is accepted}. 

\begin{figure}[h]
\vspace{-0.2cm}
    \centering
    \resizebox{0.48\textwidth}{!}{
        \includegraphics[width=9cm]{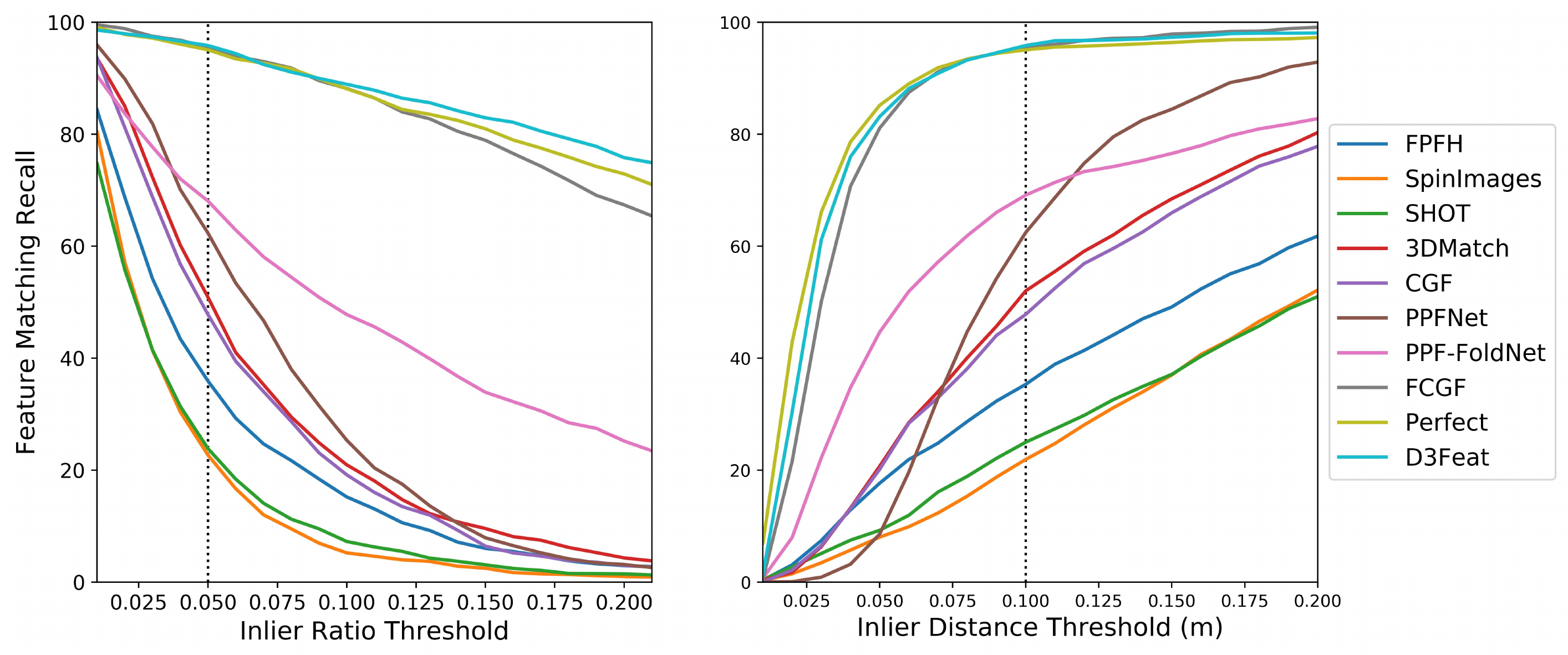}
    }
    \caption{Feature matching recall in relation to inlier ratio threshold $\tau_2$ (Left) and inlier distance threshold $\tau_1$ (Right)}
    % \hbc{In Table 1, SHOT performs much better than FPFH. Something wrong? Double check all the data here.}}
    \label{fig:recall_curve}
\end{figure}

Moreover, we demonstrate the robustness of {\Name} by varying the error threshold (originally defined as $\tau_1=10$cm and $\tau_2=5\%$) in \emph{Feature Matching Recall}. As shown in Fig.~\ref{fig:recall_curve}, we observe that the {\Name} performs consistently well under all conditions. In particular, {\Name} notably outperforms other methods (left figure) under a more strict inlier ratio threshold, e.g., $\tau_2=20\%$, which indicates the advantage of adopting a detector to derive more distinctive features. In terms of varying $\tau_1$, {\Name} performs slightly worse than PerfectMatch~\cite{gojcic2019perfect} when smaller inlier distance error is tolerated, which can be probably ascribed to the small voxel size (1.875cm) used in PerfectMatch, while {\Name} is performed using 3cm voxel downsampling.
% \hbc{Any conclusion from this figure? From this figure, seems PerfectMatch performs better than FCGF.} \\

\begin{table}[h]
% \scriptsize
    \centering
    \resizebox{0.45\textwidth}{!}{
        \begin{tabular}{l|cc|cc}
        % \hline
        \Xhline{2\arrayrulewidth}
         & \multicolumn{2}{c|}{\textbf{Origin}} & \multicolumn{2}{c}{\textbf{Rotated}} \\
         & {FMR} (\%) & {STD} & {FMR} (\%) & {STD}\\
        \hline
         \textit{FPFH~\cite{rusu2009fast}} & 35.9 & 13.4 & 36.4 & 13.6\\
         \textit{SHOT~\cite{salti2014shot}}  & 23.8 & 10.9 & 23.4 & 9.5\\
         \textit{3DMatch~\cite{zeng20173dmatch}} & 59.6 & 8.8 & 1.1 & 1.2\\
         \textit{CGF~\cite{khoury2017learning}}  & 58.2 & 14.2 & 58.5 & 14.0\\
         \textit{PPFNet~\cite{deng2018ppfnet}}   & 62.3 & 10.8 & 0.3 & 0.5\\
         \textit{PPF-FoldNet~\cite{deng2018ppf} }  & 71.8 & 10.5 & 73.1 & 10.4\\
         \textit{PerfectMatch~\cite{gojcic2019perfect}} & 94.7 & 2.7 & 94.9 & 2.5\\
         \textit{FCGF~\cite{Choy_2019_ICCV}} & 95.2 & 2.9& 95.3 & 3.3 \\
         \textit{\Name(rand)} & 95.3 & 2.7 & 95.2 & 3.2 \\
         \textit{\Name(pred)} & \textbf{95.8} & 2.9 &\textbf{ 95.5} & 3.5 \\
        % \hline
        \Xhline{2\arrayrulewidth}
        \end{tabular}
    }
    
    \caption{Feature matching recall at $\tau_1=10cm, \tau_2=5\%$. FMR and STD indicate the feature matching recall and its standard deviation.}
    \label{tab:feature_matching_recall}
\end{table}

\noindent \textbf{Performance under different number of keypoints.} To better demonstrate the superiority of a joint learning with a detector, we further report the results when reducing the sampled point number from $5000$ to 2500, 1000, 500 or even 250. As shown in Table~\ref{tab:3dmatch_num_to_perf}, when no detector is equipped, the performance of PerfectMatch, FCGF or {\NameRand} drops at a similar magnitude as the number of sampled points get smaller. However, once enabling the proposed detector ({\NamePred}), our method is able to maintain a high matching quality regarding all evaluation metrics, and outperform all comparative methods by a large margin.

It is also noteworthy that, regarding \emph{Inlier Ratio}, {\NamePred} is the only method that achieves improved results when smaller number of points is considered. This strongly indicates that the detected keypoints have been properly ranked, and the top points receive higher probability to be matched, which is a desired property that a reliable kepoint is expected to acquire.

\begin{table}[t]
	\vspace{-0.4cm}
    \centering
    \resizebox{0.48\textwidth}{!}{
        \begin{tabular}{lccccc}
         \Xhline{2\arrayrulewidth}
        \textbf{\# Keypoints} & \textbf{5000} & \textbf{2500} & \textbf{1000} & \textbf{500} & \textbf{250} \\
        \hline
            \multicolumn{6}{c}{\textit{Feature Matching Recall (\%)}} \\
        \hline
        \textit{PerfectMatch~\cite{gojcic2019perfect}} & 94.7 & 94.2 & 92.6 & 90.1 & 82.9 \\
        \textit{FCGF~\cite{Choy_2019_ICCV}}  & 95.2 & 95.5 & 94.6 & 93.0 & 89.9\\ \hdashline
        \textit{D2\_Triplet} & \multicolumn{4}{c}{Not Converge}\\
        \textit{D2\_Contrastive} & 94.7 & 94.2 & 94.0 & 93.2 & 92.3 \\ 
        \textit{w/o detector}  & 95.0 & 94.3 & 94.2 & 92.5 & 90.7 \\ \hdashline
        \textit{\NameRand} &  95.3 & 95.1 & 94.2 & 93.6 & 90.8 \\
        \textit{\NamePred} &  \textbf{95.8} &\textbf{95.6} & \textbf{94.6} & \textbf{94.3} & \textbf{93.3} \\
        \hline
          \multicolumn{6}{c}{\textit{Registration Recall (\%)}} \\
        \hline
        \textit{PerfectMatch} & 80.3 & 77.5 & 73.4 & 64.8 & 50.9 \\
        \textit{FCGF}       & \textbf{87.3} & \textbf{85.8} & \textbf{85.8} & 81.0 & 73.0\\
        \textit{\NameRand} &  83.5 & 82.1 & 81.7 & 77.6 & 68.8 \\
        \textit{\NamePred} &  82.2 & 84.4 & 84.9 & \textbf{82.5} & \textbf{79.3} \\
        \hline
          \multicolumn{6}{c}{\textit{Inlier Ratio (\%)}} \\
        \hline
        \textit{PerfectMatch} & 37.7 & 34.5 & 28.3 &  23.0 & 19.1 \\
        \textit{FCGF} & \textbf{56.9} & \textbf{54.5} & \textbf{49.1} & 43.3 & 34.7 \\
        \textit{\NameRand}  &  40.6 & 38.3 & 33.3 & 28.6 & 23.5 \\ 
        \textit{\NamePred}  & {40.7} & {40.6} & {42.7} & \textbf{44.1} & \textbf{45.0} \\
        \Xhline{2\arrayrulewidth}
        \end{tabular}
    }
    \caption{Evaluation results on the 3DMatch dataset under different numbers of keypoints.}
    \label{tab:3dmatch_num_to_perf}
\end{table}

\smallskip\noindent\textbf{Rotation invariance.} Previous works such as~\cite{deng2018ppfnet, gojcic2019perfect} use sophisticated input representations or \emph{per-point} local reference frames to acquire rotation invariance. However, as also observed in FCGF~\cite{Choy_2019_ICCV}, we find that a fully convolutional network (e.g., KPConv~\cite{thomas2019kpconv} as used in this paper) is able to empirically achieve strong rotation invariance through low-cost data augmentation, as shown in the right columns of  Table~\ref{tab:feature_matching_recall}. We will provide more details about this experiment in supplementary materials.
% \hbc{Any discussion? 3DMatch and PPFNet are not rotation-invariant? It seems all the other methods perform similarly to the original benchmark.}\xyc{Yes, 3DMatch and PPFNet are not rotation-invariant(PPF-FoldNet is rotation-invariant), And I think it is well known for readers in this area so I didn't explain i.}

\smallskip\noindent\textbf{Ablation study.} To study the effect of each component of \Name, we conduct thorough ablative experiments on 3DMatch dataset. Specifically, to address the importance of the proposed detector loss, we compare 1) the model trained with the original D2-Net loss (D2\_Triplet) and 2) the model trained with the improved D2-Net loss (D2\_Contrastive). To demonstrate the benefit from a joint learning of two tasks, we use only the contrastive loss to train a model without the detector learning (w/o detector). 
% Since point cloud fragments in 3DMatch do not have a large density variation\hbc{The reviewer might challenge you that your claim point clouds have large density variation is not the case in practice?}\xy{This is because point clouds in 3DMatch is got by fusing several point cloud scans, so the density variation is not large. But for KITTI, there are large density variations},  the effectiveness of the proposed density-invariant peakedness score is not obvious. Instead, we show the improvement in the keypoint repeatability part. 
Other training or testing settings are kept the same for a fair comparison. 

As shown in Table~\ref{tab:3dmatch_num_to_perf}, the proposed detector loss ({\NameRand}) not only guarantees better convergence than {D2\_Triplet}, but also delivers a notable performance boost over {D2\_Contrastive}. Besides, by comparing {w/o detector} and {\NameRand}, it can be seen that a joint learning of a detector is also advantageous to strengthen the descriptor itself.

\subsection{Outdoor Settings: KITTI dataset}
KITTI odometry dataset comprises 11 image sequences of outdoor driving scenarios for point cloud registration. Following the data splitting method in FCGF~\cite{Choy_2019_ICCV}, we use sequences 0 to 5 for training, 6 to 7 for validation, and 8 to 10 for testing. The ground truth transformations are provided by GPS. To reduce the noise in ground truth, we use the standard iterative closest point (ICP) method to refine the alignment.
% \hbc{This was done in the original dataset or by you?}\xyc{By me, follow FCGF paper.}
Besides, only pairs which are at least 10m away from each other are selected.
% \hbc{why?}\xyc{This is the convention for KITTI}
Finally, we train \Name~ with 30cm voxel downsampling for preprocessing.

\smallskip\noindent\textbf{Evaluation metrics.} We evaluate the estimated transformation matrices by two error metrics: Relative Translation Error (RTE) and Relative Rotation Error (RRE) proposed in~\cite{elbaz20173d, ma2016fast}. The registration is considered accurate if the RTE is below 2m and RRE is below 5$^{\circ}$ following~\cite{jian20183dfeat}.

\smallskip\noindent\textbf{Comparisions to the state-of-the-arts.} We compare {\Name} with FCGF~\cite{Choy_2019_ICCV} and 3DFeat-Net~\cite{jian20183dfeat}. For 3DFeat- Net, we report the results as presented in~\cite{Choy_2019_ICCV}. For FCGF, we use the authors' implementation trained on KITTI. To compute the transformation, we use RANSAC with 50,000 max iterations. As shown in Table~\ref{tab:kitti}, {\Name} outperforms the state-of-the-arts regarding all metrics. Besides, we also show the registration results under different numbers of points in Table~\ref{tab:kitti_num_to_reg}. In most cases, {\Name} fails to align only one pair of fragments.
% \hbc{Why compared to FCGF only?}\xyc{Only FCGF and 3DFeatNet have been tested on KITTI and 3DFeatNet is much worse}. 
Even using only 250 points, {\Name} still achieves 99.63\% success rate, while FCGF drops to 68.62\% due to the lack of a reliable detector.
% \hbc{why not test the performance under 500 and 250 keypoints?}.
% Noticed that all the previous methods use all the points in the fragment to build correpondence set and do the RANSAC, while our method utilize only 5000 keypoints predicted by our detector network.

\begin{table}[ht]
    \centering
    \resizebox{0.48\textwidth}{!}{
        \begin{tabular}{l|cc|cc|c}
        \Xhline{2\arrayrulewidth}
        & \multicolumn{2}{c|}{\textbf{RTE(cm)}} & \multicolumn{2}{c|}{\textbf{RRE($^{\circ}$)}} & \textbf{Succ.(\%)}\\
        & AVG & STD & AVG & STD & \\
        % \textbf{Method} & \textbf{RTE(cm)} & \textbf{STD} & \textbf{RRE()} & \textbf{STD} & \textbf{Succ.(\%)}  \\
        \hline
        \textit{3DFeat-Net~\cite{jian20183dfeat}} & 25.9 & 26.2 & 0.57 & 0.46 & 95.97 \\
       	\textit{FCGF~\cite{Choy_2019_ICCV}} & 9.52 & 1.30 & 0.30 & 0.28 & 96.57 \\
        \textit{\Name} & \textbf{6.90} & \textbf{0.30} & \textbf{0.24} & \textbf{0.06} & \textbf{99.81} \\
        % Ours(20cm) & & & & &\\
         \Xhline{2\arrayrulewidth}
        \end{tabular}
    }
    \caption{Quantitative comparisons on the KITTI dataset. The results of 3DFeat-Net are taken from~\cite{Choy_2019_ICCV}. {For RTE and RRE, the lower of the values, the better.}
    %For FCGF, We use the origin implementation from the authors but run RANSAC for only 50000 max iteration and 1000 max validation. 
    }
    \label{tab:kitti}
\end{table}

\begin{table}[ht]
	\vspace{-0.4cm}
    \centering
    \resizebox{0.48\textwidth}{!}{
    \begin{tabular}{l|cccccc}
     \Xhline{2\arrayrulewidth}
    \textbf{} & \textbf{All} & \textbf{5000} & \textbf{2500} & \textbf{1000} & \textbf{500} & \textbf{250}   \\
    \hline
   \textit{FCGF~\cite{Choy_2019_ICCV}} & 96.57 & 96.39 & 96.03 & 93.69 & 90.09 & 68.62\\
    \textit{\Name} & 99.81 & 99.81 & 99.81 & 99.81 & 99.81 & 99.63\\
     \Xhline{2\arrayrulewidth}
    \end{tabular}
    }
    \caption{Success rate on the KITTI dataset under different numbers of keypoints. }
    \label{tab:kitti_num_to_reg}
\end{table}

\subsection{Outdoor settings: ETH Dataset}
% \hbc{This dataset is not mentioned in the Introduction.} 

In order to evaluate the generalization ability of \Name, we use {the} model trained on 3DMatch dataset and test {it} on four outdoor laser scan datasets (Gazebo-Summer, Gazebo-Winter, Wood-Summer and Wood-Autumn) from {the} ETH dataset~\cite{Pomerleau:2012}, following the protocols defined in~\cite{gojcic2019perfect}. The evaluation metric is again \emph{Feature Matching Recall} under the same settings {as} in previous evaluations. For comparisons, we use the PerfectMatch model with 6.25cm voxel size (16 voxels per 1m), while for FCGF, we use the model trained on 5cm voxel size which we find perform significantly better than the model with 2.5cm on this dataset. For \Name, we report the results on both 6.25cm and 5cm to compare with PerfectMatch and FCGF, respectively.
\begin{table}[ht]
\vspace{-0.1cm}
    \centering
    \resizebox{0.48\textwidth}{!}{
        \begin{tabular}{l|c|cccc|c}
        \Xhline{2\arrayrulewidth}
        & voxel & \multicolumn{2}{c}{\textbf{Gazebo}} & \multicolumn{2}{c|}{\textbf{Wood}} & \textbf{}\\
        & size(cm) & Sum. & Wint. & Sum. & Aut. & AVG. \\
        % \textbf{Method} & \textbf{RTE(cm)} & \textbf{STD} & \textbf{RRE()} & \textbf{STD} & \textbf{Succ.(\%)}  \\
        \hline
        \textit{PerfectMatch~\cite{gojcic2019perfect}}  & 6.25 & \textbf{91.3} & \textbf{84.1} & \textbf{67.8} & \textbf{72.8} & \textbf{79.0}\\ 
        \textit{FCGF~\cite{Choy_2019_ICCV}}  & 5.00 &  22.8 & 10.0 & 14.78 & 16.8 & 16.1 \\
        \textit{\NameRand}& 5.00 & 45.7 & 23.9 & 13.0 & 22.4 & 26.2 \\
        \textit{\NamePred}& 5.00 & 78.9 & 62.6 & 45.2 & 37.6 & 56.3 \\
        \textit{\NamePred}& 6.25 & 85.9 & 63.0 & 49.6 & 48.0 & 61.6\\
         \Xhline{2\arrayrulewidth}
        \end{tabular}
    }
    \caption{Feature matching recall at $\tau_1=10cm$, $\tau_2=5\%$ on the ETH dataset.}
    \label{tab:eth}
\end{table}

% PerfectMatch achieves satisfactory results but both Ours and FCGF cannot generalize well on the ETH dataset\hbc{This doesn't sound good. Maybe presenting this part as one of the limitations of our work?}.
For a fair comparison, the pre-sampled points provided in~\cite{gojcic2019perfect} are used for PerfectMatch, FCGF and {\NameRand} to extract the features. As shown in Table~\ref{tab:eth}, under 5cm voxel size setting, {\Name} demonstrates better generalization ability than FCGF. Once the detector is enabled, the results are improved remarkably. However, on this dataset, PerfectMatch is still the best performing method, whose generalization ability can be mainly ascribed to the use of smoothed density value (SDV) representation, 
%\xy{which reduces the density difference of the input point clouds even captured from a scenario different from training~\cite{gojcic2019perfect}.}
 as explained in the original paper~\cite{gojcic2019perfect}. 
% The generalization ability of PerfectMatch mainly comes from their proposed smoothed density value (SDV) representation. SDV performs smoothing to an input point cloud during the construction of input patches, thus greatly reducing the sparsity change of the point cloud.}
% \hb{It is mainly because} point cloud scans in the ETH dataset mainly contain sparse and dense \hbc{how come they are both sparse and dense?} vegetation like trees and bushes, which is very different from \hb{the indoor 3DMatch benchmark}, where most objects in the scenes have flat surfaces. 
% The SDV representation used by PerfectMatch performs smoothing to \hb{an input} point cloud during the construction of input patches, \hb{thus greatly reducing} the sparsity of the input point cloud. Thanks to such \hb{a} smoothing mechanism, PerfectMatch \hb{is} less sensitive to the sparsity changes of the point cloud\hbc{But you claimed our method is sparsity invariant? The discussions here are very confusing, since it is somewhat in conflict with the claims you made before?}, \hb{and} thus generalizes well from indoor scenes to outdoor scenes. While FCGF and our model have no mechanism to perform such completion\hbc{Can such a mechanism be integrated with our approach?}. This experiment also \hb{implies} a potential future direction, i.e., to implicitly perform point cloud smoothing \hbc{Is this difficult to do?} in the network. 
Nevertheless, by comparing the results of {\NameRand} and {\NamePred}, we can still find the significant improvement brought by our detector. The visualization results on ETH can be found in the supplementary materials.
 
\subsection{Keypoint Repeatability}
    \label{section:Exp_keypoint}
 Besides the reliability, a keypoint is also desired to be repeatable.
%  another desired property of selected keypoints is repeatability and accurate localization under arbitrary transformations. 
 Thus we compare our detector {in} repeatability with a learning-based method USIP~\cite{li2019usip} and hand-crafted methods: ISS~\cite{zhong2009intrinsic},  Harris-3D~\cite{harris1988combined}, and SIFT-3D~\cite{lowe2004distinctive} on the 3DMatch test set and KITTI test set. 
 
\smallskip\noindent\textbf{Evaluation metric.} We use {the} relative repeatability proposed in~\cite{li2019usip} as the evaluation metric. Specifically, given two partially overlapped point clouds and the ground truth pose, a keypoint in the first point cloud is {considered} repeatable if its distance to the nearest keypoint in the second point cloud is less than some threshold, under the ground truth transformation. Next, the relative repeatability is calculated {as} the number of repeatable keypoints over the number of detected keypoints. This threshold is set to 0.1m and 0.5m for 3DMatch and  KITTI datasets, respectively.
  
\smallskip\noindent \textbf{Implementation.} We compare the full model \Name, and a baseline model that directly {extends the} D2-Net design to 3D domain (denoted as \Name(base)). We use PCL~\cite{pcl} to implement the classical detectors. For USIP, we take the origin implementation as well as the pretrained model to test on the 3DMatch test set. For the KITTI dataset, since USIP and \Name~ use different processing and splitting {strategies} and USIP requires surface normal and curvature as input, the results are not directly comparable. Nevertheless, for the sake of completeness, we run {and test} the original implementation of USIP on our processed KITTI {data}. %test set and report the result. 
For each detector, we generate 4, 8, 16, 32, 64, 128, 256, 512 keypoints and calculate the relative repeatability respectively. 
 
\smallskip\noindent\textbf{Ablation study.} To recap, direct extension from D2-Net to 3D domain brings the problem that points located in more sparse region would have higher probability of being selected as keypoints, which prevents the network from predicting highly repeatable keypoints, as explained in Sec.~\ref{section:methodology}. The proposed density-invariant selection strategy enables the network to handle the inherent density variations of point clouds. In the following, we will demonstrate the effect of proposed keypoint selection strategy through both qualitative and quantitative results.

\begin{figure}[h]
 \vspace{-0.3cm}
    \centering	 
    \includegraphics[width=8.5cm]{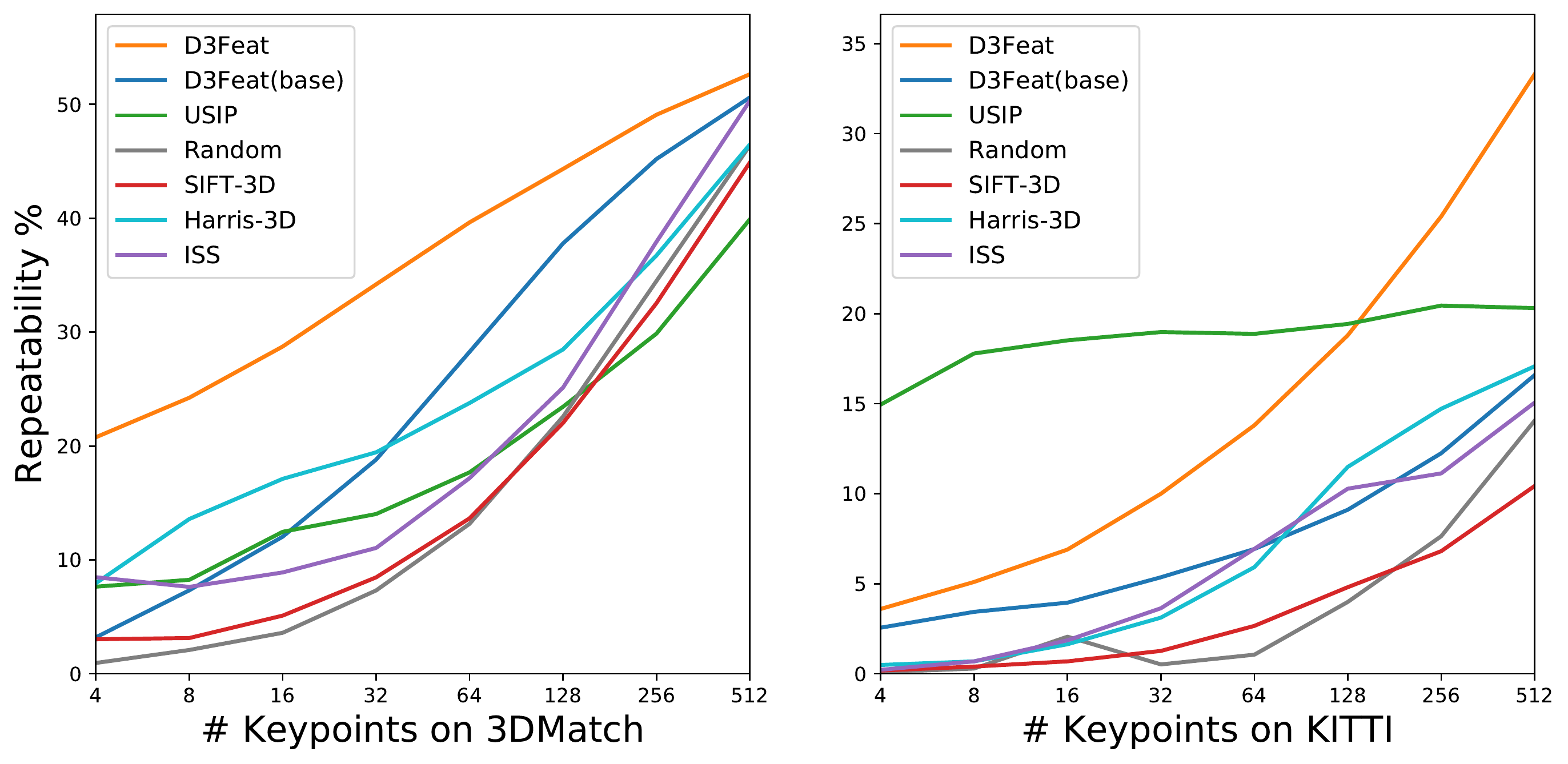}
    \caption{Relative repeatability when different numbers of keypoints are detected.}
    \label{fig:repeat_curve}
\end{figure}

\smallskip\noindent \textbf{Qualitative results.} The relative repeatability in relation to the number of keypoints is shown in Fig.~\ref{fig:repeat_curve}. Thanks to the proposed self-supervised detector loss, our detector is encouraged to give {higher scores} to points with distinctive local geometry while {giving lower scores} to points with homogeneous neighborhood. Therefore although we do not explicitly supervise the repeatability of keypoints, {\Name} can still achieve comparable repeatability {to USIP}. {\Name} generally
outperforms all the other detectors {on 3DMatch and KITTI dataset over all different number of keypoints except worse than USIP on KITTI}. In addition, the proposed saliency score significantly improves the repeatability of {\Name} over \Name(base),
%(see the orange curve and blue curve),
{indicating the effectiveness of the proposed keypoint selection strategy.}
% \tai{nothing specific to elaborate here?}
 
\begin{figure}[h]
    \vspace{-0.3cm}
    \centering
    \includegraphics[width=9cm]{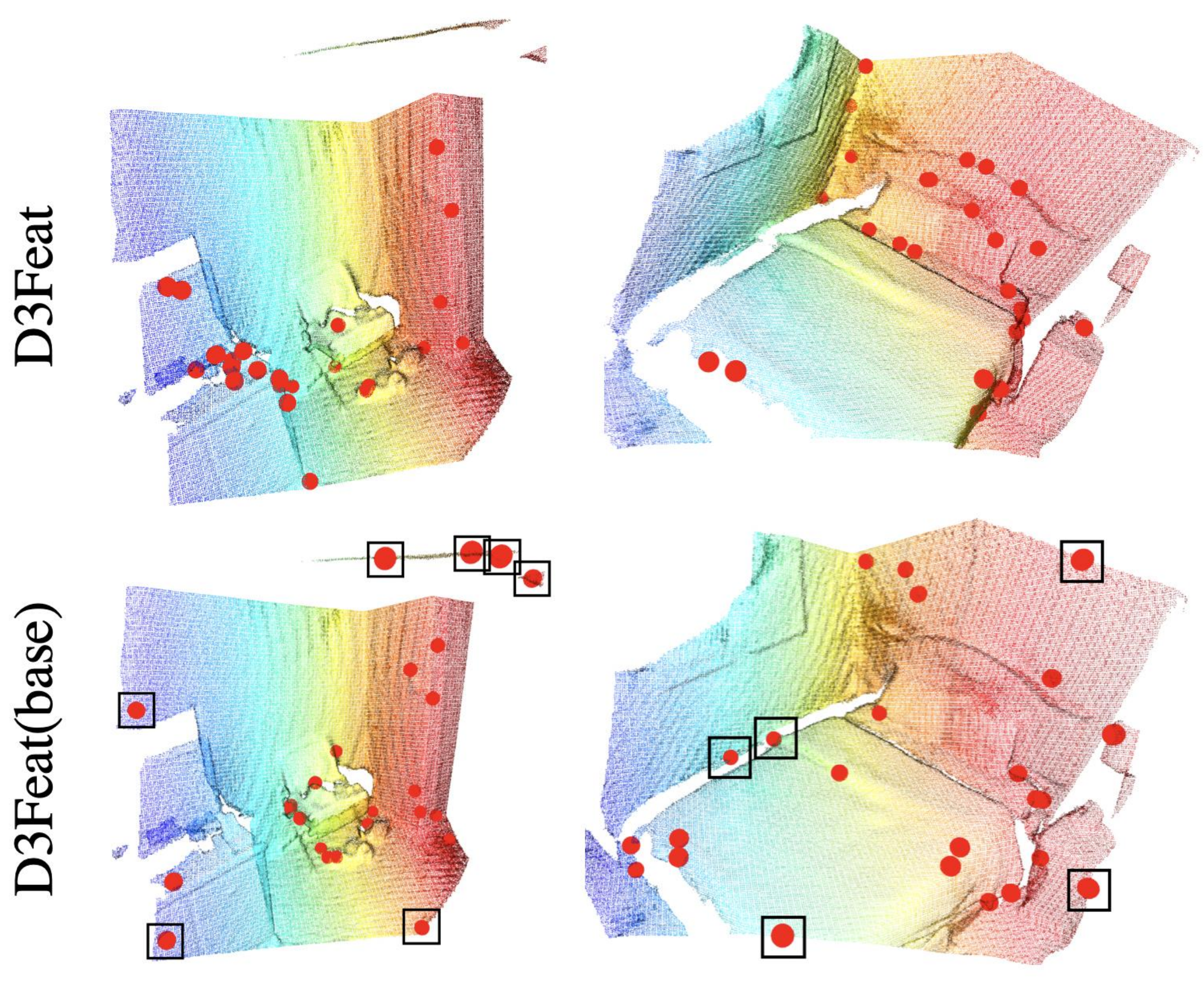}
    \caption{Visualization on 3DMatch. The first row are detected using D3Feat while the second row are detected using na\"{i}ve local max score. Points close to the boundaries are marked with black boxes.}
    \label{fig:3dmatch_keypts}
    \end{figure}

\begin{figure}[h]
    \vspace{-0.1cm}
    \centering
    \includegraphics[width=9cm]{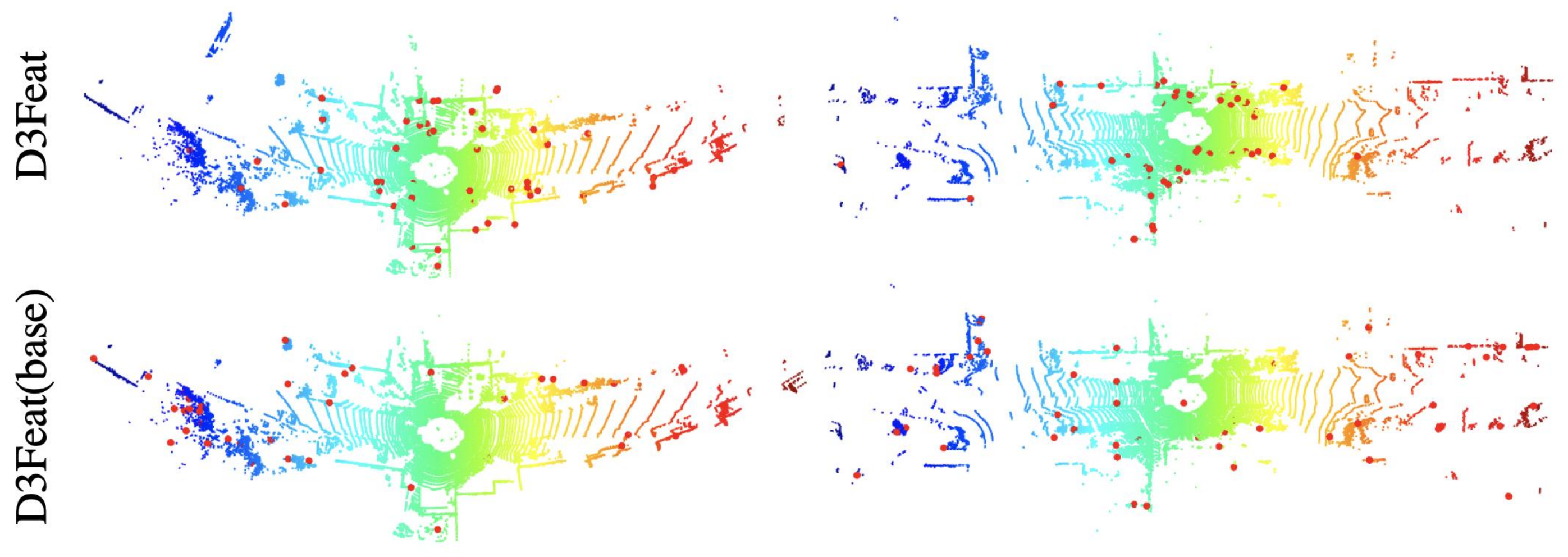}
    \caption{Visualization on KITTI. The first row are detected using D3Feat while the second row are detected using na\"{i}ve local max score. Best view with color and zoom-in.}
    \label{fig:ablation_peakedness}
    \end{figure}

 \noindent \textbf{Qualitative results.} 
 %Fig.~\ref{fig:ablation_peakedness} and Fig.~\ref{fig:3dmatch_keypts} show {the visualization of representative results of detected keypoints} on {KITTI and 3DMatch, respectively}. 
 %Directly extending the D2-Net design to the 3D domain brings the problem that points located in more sparse areas would have higher probability of being selected as keypoints, as explained in Sec.~\ref{section:methodology}.  
As shown in Fig.~\ref{fig:3dmatch_keypts} and Fig.~\ref{fig:ablation_peakedness}, the proposed density-invariant selection strategy overcomes the density variations. More qualitative results are included in the supplementary materials. 

%% file: Conclusion.tex
\vspace{-0.2cm}
In this paper we have designed a dual-role fully convolutional network for dense feature detection and description on point clouds. A novel density-invariant saliency score has been proposed to help select keypoints under varying densities. Also a self-supervised detector loss has been proposed to guide the network to predict highly repeatable keypoints and enable joint improvement for both {detector} and descriptor. {Extensive} experiments on indoor 3DMatch and outdoor KITTI both show the effectiveness of our detector network and the distinctiveness of descriptor network. Our model outperforms the state-of-the-art methods especially when using a small number of keypoints. \\% Our model also shows strong generalizability on ETH dataset. \\

\noindent\textbf{Acknowledgement.} This work is supported by Hong Kong RGC GRF (No.~16206819 and No.~16203518) and Centre for Applied Computing and Interactive Media (ACIM) of School of Creative Media, City University of Hong Kong.
% \hbc{No discussions on the limitations and future work?} \xyc{will add future work discussion in supplementary material.}

%% file: Supplementary.tex
In this supplementary material, we first explain the details about our modification on KPConv~\cite{thomas2019kpconv}, then we analyze the contribution of rotation augmentation by an ablation study on 3DRotatedMatch dataset. {We further conduct the ablative experiments on the detector loss as well as incorporating our detector with FCGF.} Finally, we provide additional details about the experiments on 3DMatch, KITTI and ETH datasets and some further visualization results.

\subsection{Implementation Details}
\noindent\textbf{Normalization term} As explained in the main paper, the original formulation of KPConv is not invariant to point density. Thus we add a density normalization term, which sums up the number of the supporting points in the neighborhood, to ensure that the convolution is sparsity invariant. To demonstrate the effectiveness of the normalization term, we train networks with and without the normalization term in the same settings with what is described in the main paper, and report the registration results on 3DMatch dataset. 
% The point cloud fragments in 3DMatch dataset are very dense, thus using voxel downsample results in fragments with almost uniform density.
During testing, instead of using voxel downsample, we use {uniform downsample} i.e. \texttt{uniform\_down\_sample} in Open3D~\cite{zhou2018open3d} implementation, by which the density variations of input point clouds is enlarged. We evaluate the model on sample rate $=15$, which leads to a similar average number of points for the point clouds in the test set. The result is shown in Table~\ref{tab:ablation_norm}.

\begin{table}[h]
% \scriptsize
    \centering
    \resizebox{0.48\textwidth}{!}{
        \begin{tabular}{c|c|ccccc}
        % \hline
        \Xhline{2\arrayrulewidth}
         & \multicolumn{1}{c|}{\textbf{Voxel}} & \multicolumn{5}{c}{\textbf{Uniform}} \\
        & 5000 & 5000 & 2500 & 1000 & 500 & 250\\
        \hline
         \textit{w/o norm.}  & 95.6 & 77.3 & 76.5 & 73.9 & 71.4 & 66.1 \\
         \textit{w norm.} & 95.8 & 91.9 & 91.8 & 91.2 & 90.2 & 89.4  \\
        % \hline
        \Xhline{2\arrayrulewidth}
        \end{tabular}
    }
    \caption{Feature matching recall at $\tau_1=10cm, \tau_2=5\%$. \textbf{Voxel} indicates voxel downsample with voxel size = $0.03m$, while \textbf{Uniform} indicates uniform downsample with sample rate = $15$.}
    \label{tab:ablation_norm}
\end{table}

As shown in Table~\ref{tab:ablation_norm}, the normalization term is effective to handle neighborhoods with different sparsity levels. When using a uniform downsample strategy, {\Name} with the normalization term is able to maintain a high matching quality, and outperform the model without the normalization term by a large margin.\\

\noindent\textbf{Network architecture} We adopt the architecture for segmentation tasks proposed in KPConv~\cite{thomas2019kpconv}. The number of channels in the encoder part are $(64, 128, 256, 512, 1024)$. Skip connections are used between the corresponding layers of the encoder part and the decoder part. The output features are processed by a 1$\times$1 convolution to get the final 32 dimensional features. 
% For the parameters of convolution layers, each layer $j$ has a cell size $dl_j$ from which other parameters can be inferred. The neighborhood radius is $2.5dl_j$ and average kernel radius is $1.5dl_j$. The kernel points influence distance is set as $1.0dl_j$ in the original paper, while we find that setting it as $2.0dl_j$ gives better performance.
Other settings are the same with original paper~\cite{thomas2019kpconv}.

\subsection{Ablation on Rotation Invariance}
In experiments, we find that a fully convolutional network is able to empirically achieve strong rotation invariance through low cost data augmentation.  To demonstrate the effectiveness of  simple data augmentation on the robustness to rotation, we further train the model without rotation augmentation and evaluate it on 3DRotatedMatch dataset.  The result is shown in Table~\ref{tab:ablation_aug}. 

\begin{table}[h]
    \centering
    \resizebox{0.40\textwidth}{!}{
    \begin{tabular}{c|ccccc}
     \Xhline{1.5\arrayrulewidth}
    \textbf{} & \textbf{5000} & \textbf{2500} & \textbf{1000} & \textbf{500} & \textbf{250}   \\
    \hline
    \textit{w/o aug.} & 5.5 & 5.4 & 4.9 & 5.4 & 5.5\\
    \textit{w aug.}  & 95.5 & 95.0 & 94.8 & 92.1 & 88.9\\
     \Xhline{1.5\arrayrulewidth}
    \end{tabular}
    }
    \caption{Feature match recall on 3DRotatedMatch with and without rotation augmentation.}    \label{tab:ablation_aug}
\end{table}

Without rotation augmentation, {\Name} fails on 3DRotatedMatch because the network cannot learn the rotation invariance from the data. \\

\subsection{Ablation on Detector Loss}

To better analyze the contribution of the proposed detector loss, we re-create a table below, which derives original results from Table~\ref{tab:3dmatch_num_to_perf}, and the results from our model without the detector loss and performing on the predicted keypoints(w/o detector~(pred)). The effect of detector loss can be seen from the comparison between {w/o detector} and {\Name} on both random (\emph{rand}) or predicted (\emph{pred)} points. It is shown that the detector loss not only strengthens the descriptor itself (given random keypoints), but also boosts the performance of the detector (given predicted keypoints).
It is also noteworthy that only {\NamePred} improves the \textit{Inlier Ratio} when reducing the number of points, which clearly indicates that the detector loss helps to better rank the keypoints regarding distinctiveness.

\begin{table}[h]
	\vspace{-0.2cm}
    \centering
    \resizebox{0.48\textwidth}{!}{
        \begin{tabular}{lccccc}
         \Xhline{2\arrayrulewidth}
        \textbf{\# Keypoints} & \textbf{5000} & \textbf{2500} & \textbf{1000} & \textbf{500} & \textbf{250} \\
        \hline
            \multicolumn{6}{c}{\textit{Feature Matching Recall (\%)}} \\
        \hline
        \textit{w/o detector(rand)}  & 95.0 & 94.3 & 94.2 & 92.5 & 90.7 \\ 
       \textit{w/o detector(pred)}  & 95.2 & 95.0 & 94.5 & 92.4 & 90.4 \\ \hdashline
        \textit{\NameRand} &  95.3 & 95.1 & 94.2 & 93.6 & 90.8 \\
        \textit{\NamePred} &  \textbf{95.8} &\textbf{95.6} & \textbf{94.6} & \textbf{94.3} & \textbf{93.3} \\
        \hline
          \multicolumn{6}{c}{\textit{Inlier Ratio (\%)}} \\
        \hline
        \textit{w/o detector(rand)}  & 40.7 & 39.9 & 35.2 & 31.0 & 25.1 \\ 
        \textit{w/o detector(pred)}  & \textbf{41.7} & 40.3 & 38.7 & 36.6 & 33.2 \\ \hdashline
        \textit{\NameRand}  &  40.6 & 38.3 & 33.3 & 28.6 & 23.5 \\ 
        \textit{\NamePred}  & {40.7} & \textbf{40.6} & \textbf{42.7} & \textbf{44.1} & \textbf{45.0} \\
        \Xhline{2\arrayrulewidth}
        \end{tabular}
    }
    \caption{Ablation study on the proposed detector loss.}
    \label{tab:ablation_detectorloss}
\end{table}

\subsection{Ablation on Detector with FCGF}

Since the detection scores are computed on top of the extracted dense descriptors, it is easy to incorporate our detector with other dense feature description models, such as FCGF~\cite{Choy_2019_ICCV}. To demonstrate the usability of our method, we train the FCGF with the proposed joint learning method  and evaluate it on the 3DMatch dataset, as shown in Table~\ref{tab:fcgf_detector}. The model FCGF + detector is trained using the proposed detector loss under the same setting with \cite{Choy_2019_ICCV} for 100 epochs.
\begin{table}[h]
	\vspace{-0.2cm}
    \centering
    \resizebox{0.48\textwidth}{!}{
        \begin{tabular}{lccccc}
         \Xhline{2\arrayrulewidth}
        \textbf{\# Keypoints} & \textbf{5000} & \textbf{2500} & \textbf{1000} & \textbf{500} & \textbf{250} \\
        \hline
            \multicolumn{6}{c}{\textit{Registration Recall (\%)}} \\
        \hline
        \textit{FCGF\cite{Choy_2019_ICCV}}       & \textbf{87.3} & {85.8} & {85.8} & 81.0 & 73.0\\
       \textit{FCGF + detector}  & 86.7 & \textbf{87.8} & \textbf{88.3} & \textbf{85.4} & \textbf{81.5} \\

        \hline
          \multicolumn{6}{c}{\textit{Inlier Ratio (\%)}} \\
        \hline
      \textit{FCGF\cite{Choy_2019_ICCV}} & \textbf{56.9} & \textbf{54.5} & {49.1} & 43.3 & 34.7 \\
        \textit{FCGF + detector}  &  53.5 & 53.2 & \textbf{53.6} & \textbf{53.2} & \textbf{51.0} \\ 
        \Xhline{2\arrayrulewidth}
        \end{tabular}
    }
    \caption{Evaluation results of FCGF trained with the proposed detector loss.}
    \label{tab:fcgf_detector}
\end{table}

The result shows that FCGF can indeed benefit from the proposed joint learning and maintain a high performance given a smaller number of points.

\subsection{Runtime}

To demonstrate the efficiency of our method, we compare the runtime of {\Name} with FCGF~\cite{Choy_2019_ICCV} on 3DMatch in Table~\ref{tab:runtime}. For a fair comparison, we use the same voxel size (2.5cm, roughly 20k points) with FCGF to measure the runtime of {\Name} including both detection and description. The performance difference mainly lies in the sparse convolution used by FCGF, which is time-consuming in hashing.

\begin{table}[ht]
    \centering
    \resizebox{0.48\textwidth}{!}{
    \begin{tabular}{l|cccc}
     \Xhline{2\arrayrulewidth}
    \textbf{} & \textbf{CPU} & \textbf{GPU}  & \textbf{Time(s)}  \\
    \hline
    \textit{FCGF\cite{Choy_2019_ICCV}} & Intel 10-core 3.0GHz(i7-6950) & Titan-X & 0.36\\
    \textit{\Name} & ~Intel 4-core 4.0GHz(i7-4790K) & GTX1080 & \textbf{0.13}\\
     \Xhline{2\arrayrulewidth}
    \end{tabular}
    }
    \caption{Average runtime per fragment on 3DMatch test set.}
    \label{tab:runtime}
\end{table}

\subsection{Evaluation Metric for 3DMatch}
%This section provides a detailed explanation of the evaluation metrics on 3DMatch benchmark including feature matching recall and registration recall.\\
\noindent\textbf{Feature matching recall} Feature matching recall is first proposed in~\cite{deng2018ppfnet}, which measures the quality of features without using a RANSAC pipeline. Given two partially overlapped point cloud $P$ and $Q$, and the descriptor network denoted as a non-linear function $f$ mapping from input points to feature descriptors, the correspondence set for the fragments pairs is obtained by mutually nearest neighbor search in feature space,
    \begin{equation}
        \begin{aligned}
        \Omega = \{p_i \in P, q_j \in Q | & f(p_i) = nn(f(q_j), f(P)), \\
        & f(q_j) = nn(f(p_i), f(Q))  \},
        \end{aligned}
    \end{equation}
where $nn()$ denotes the nearest neighbor search based on the Euclidean distance. Finally the feature matching recall is defined as,
\begin{equation}
    R = \frac{1}{|M|} \sum_{m=1}^{|M|} \mathds{1}\Big( \Big[\frac{1}{|\Omega|} \sum_{(i,j)\in \Omega} \mathds{1}(||p_i - T_mq_j|| < \tau_1)\Big] > \tau_2 \Big) 
\end{equation}
where $M$ is the set of point cloud fragment pairs which have more than 30\% overlap, and $T_{m}$ is the ground truth transformation between the fragment pair $m \in M$. $\tau_1$ is the inlier distance threshold between a correspondence pair,  and $\tau_2$ is the inlier ratio threshold of the fragment pair. Following the setting of~\cite{zeng20173dmatch}, the correspondence which have less than $\tau_1=10cm$ euclidean distance between their descriptors are seen as inliers, and the fragment pairs which have more than $\tau_2=5\%$ inlier correspondences will be counted as one match. The evaluation metric is based on the theoretical analysis that RANSAC need $k = 55258$ iterations to achieve 99.9\% confidence of finding at least 3 correspondence with inlier ratio 5\%. \\

\noindent\textbf{Registration recall}  Registration recall~\cite{zeng20173dmatch} measures the quality of features within a reconstruction system, which firstly uses a robust local registration algorithm like RANSAC to estimate the rigid transformation between two point clouds, then calculate the RMSE of the ground truth correspondence under the estimated transformation. The ground truth correspondence set for fragments pair $P$ and $Q$ is given,
    \begin{equation}
        \Omega^* = \{ p^* \in P, q^* \in Q\}
    \end{equation}
then the registration recall is defined as,
    \begin{equation}
        R = \frac{1}{|M|} \sum_{m=1}^{|M|} \mathds{1}\Big( \sqrt{\frac{1}{|\Omega^*|}\sum_{(p^*,q^*) \in \Omega^*} ||p^* - \hat T_mq^* ||^2} < 0.2 \Big),
    \end{equation}
    where $\hat T$ is the transformation matrix estimated by RANSAC. In our experiment, we run a maximum of 50,000 iterations on the initial correspondence set to estimate the transformation between fragments following~\cite{zeng20173dmatch}.
    
\subsection{Dataset Preprocessing}
This section provides the steps to process the datasets including 3DMatch, 3DRotatedMatch, KITTI and ETH.\\

\noindent\textbf{3DMatch} For training set, we follow the steps in~\cite{zeng20173dmatch} to get fused point cloud fragments and corresponding poses. We find all the fragments pairs that have more than $30\%$ overlap to build the training set. During training, we alternate between selecting the nearby fragment as the corresponding pair, or randomly selecting from all the overlapped fragments for fast convergence. For test set, we directly use the point cloud fragments and ground truth poses provided by the authors without performing any preprocess to extract the dense feature and score map. \\

\noindent\textbf{3DRotatedMatch} Our model is inherently translation invariant because we are using the relative coordinates. So in order to test the robustness of our model to rotation, we create the 3DRotatedMatch test set following~\cite{deng2018ppf}. We rotate all the fragments in 3DMatch test set along all three axes with random sampled angle from a uniform distribution over [0, 2$\pi$).\\

\noindent\textbf{KITTI} The training set of KITTI odometry dataset contains 11 sequences, we use sequence 0 to 5 for training, sequence 6 to 7 for validation and the last three for testing. Since GPS ground truth is noisy, we first use ICP to refine the alignment and then verify by whether enough correspondence pairs can be found. We select Lidar scan pairs with at least 10m intervals to obtain 1358 pairs for training, 180 pairs for validation and 555 for testing. \\

\noindent\textbf{ETH} For a fair comparison, we directly use the raw point clouds, the ground-truth transformations along with the overlap ratio provided by the authors of \cite{gojcic2019perfect} to extract the features and evaluate the registration results. 

\subsection{Qualitative Visualization}
\begin{figure*}[ht]
\vspace{-0.3cm}
    \centering
    \includegraphics[width=17cm]{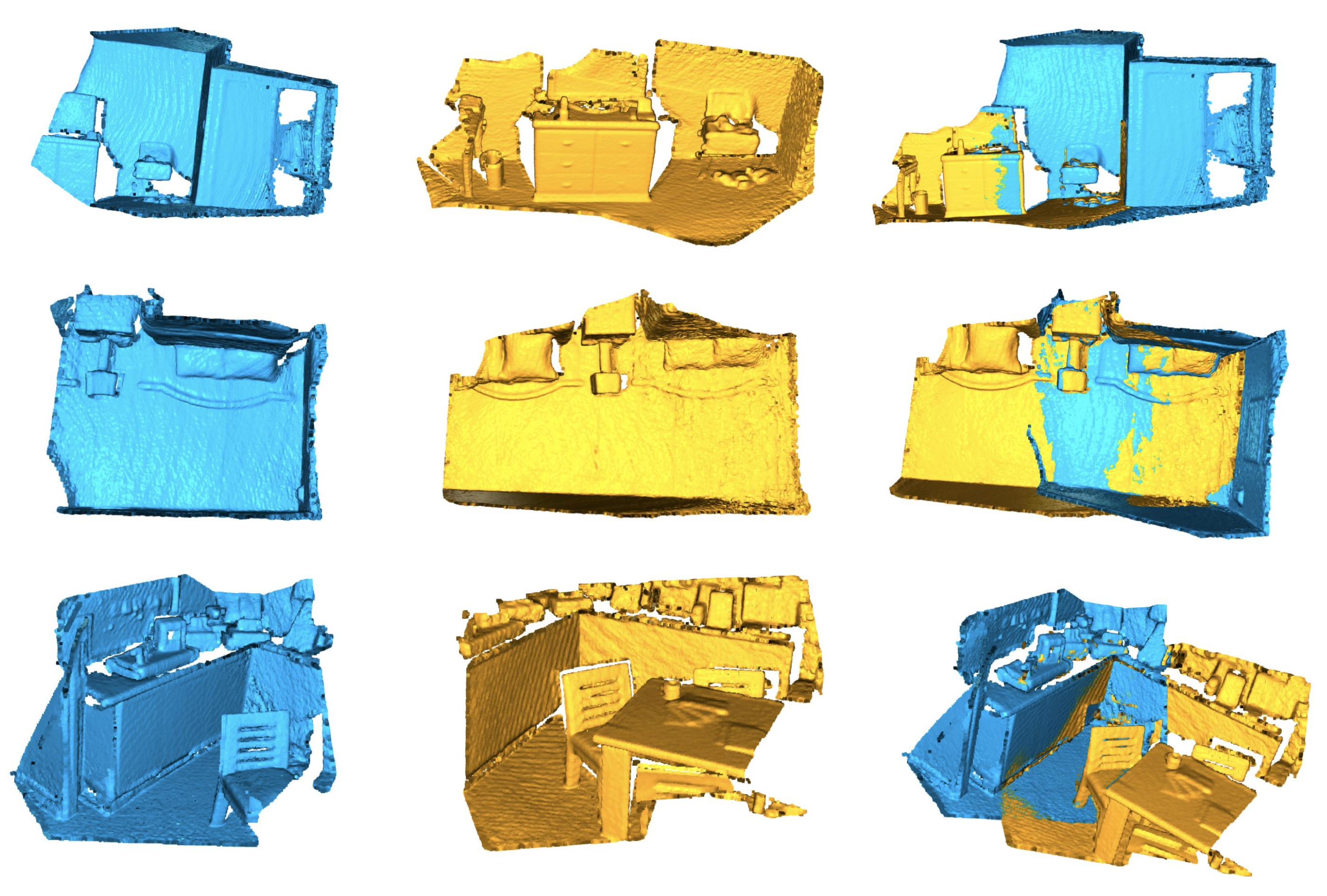}
    \\
    \includegraphics[width=17cm]{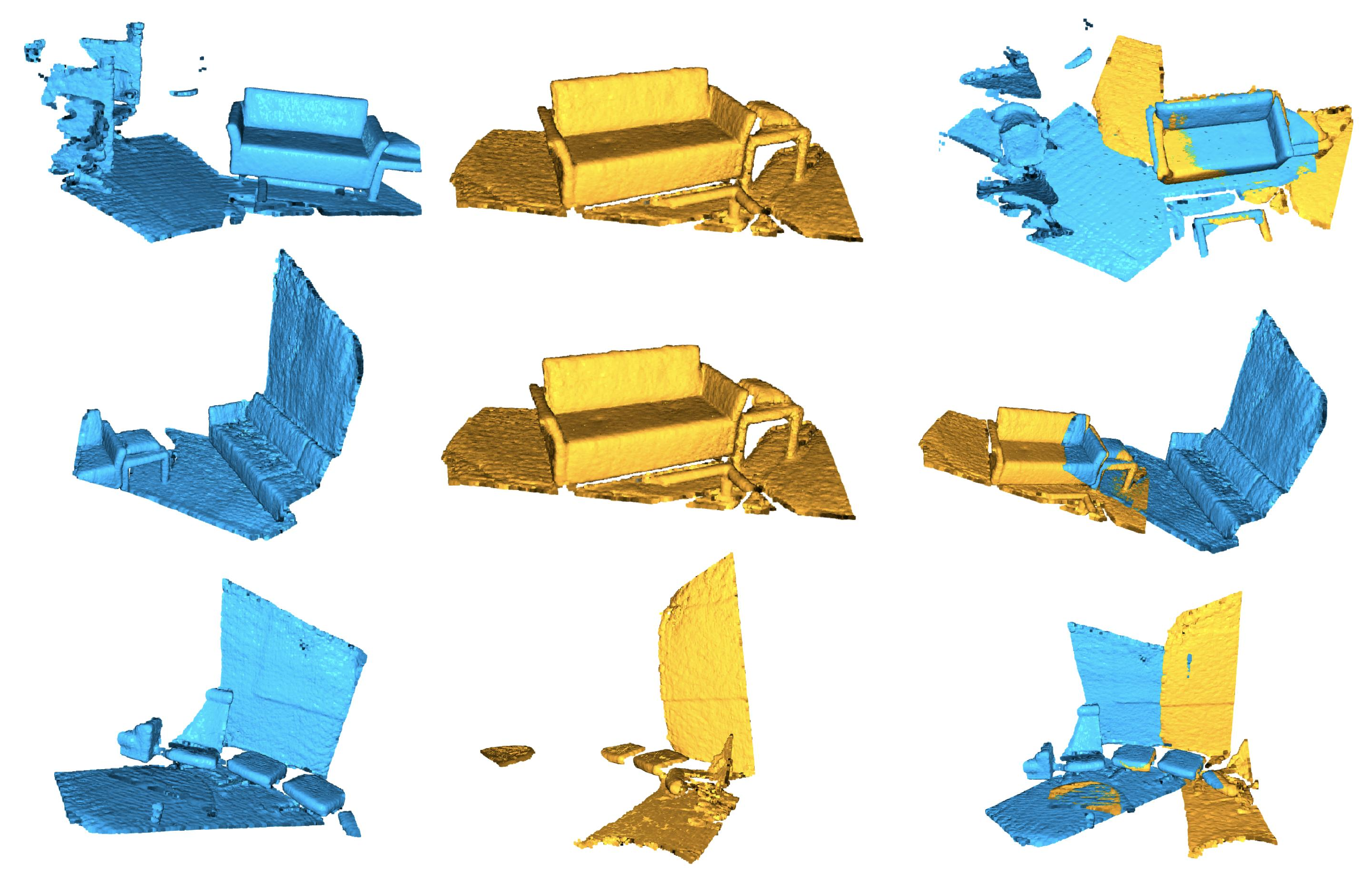}
     \caption{Qualitative results on the 3DMatch dataset. The first two columns are input point cloud fragments, and the third column presents the registration results. Best view with color and zoom-in.}
    \label{fig:vis_registration}
\end{figure*}

\begin{figure*}[ht]
    \centering
    \includegraphics[width=18cm]{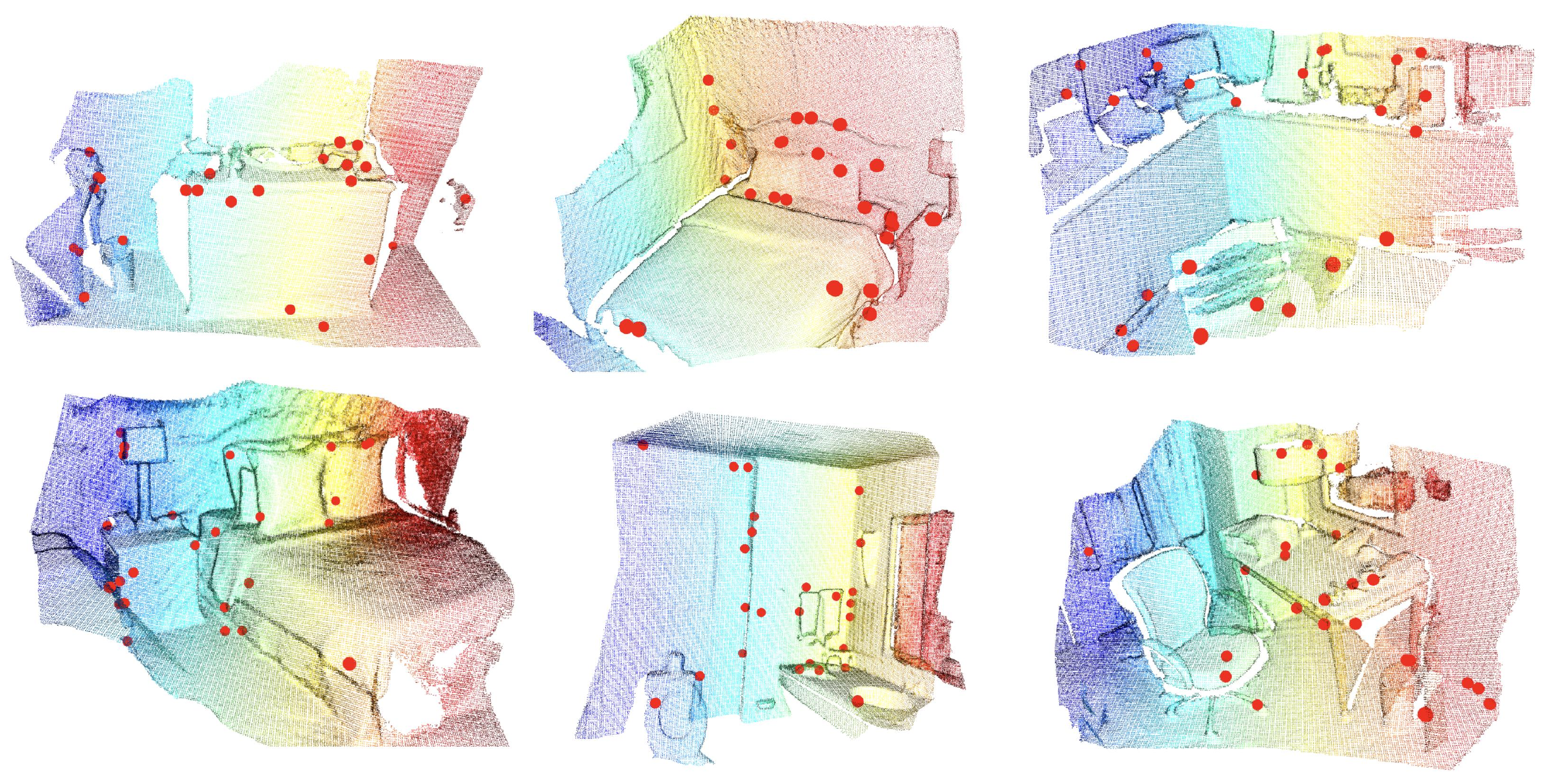}
     \caption{Visualization of keypoints on the 3DMatch dataset. Best view with color and zoom-in.}
    \label{fig:vis_3dmatch}
\end{figure*}

\begin{figure*}
    \centering
    \includegraphics[width=18cm]{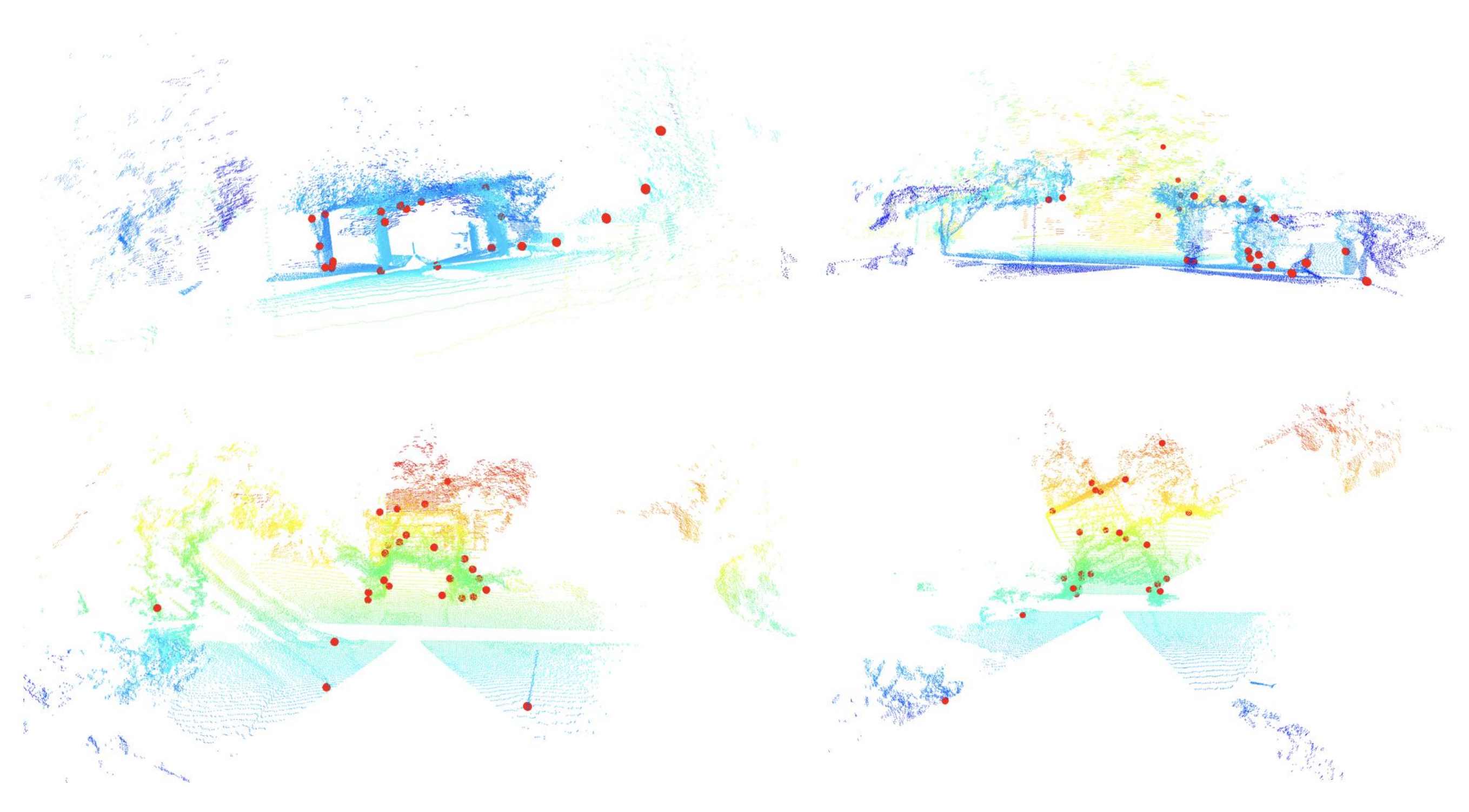}
    \caption{Visualization of keypoints on ETH dataset. Best view with color and zoom-in.}
    \label{fig:vis_eth}
\end{figure*}

\begin{figure*}[h]
    \centering
    \includegraphics[width=18cm]{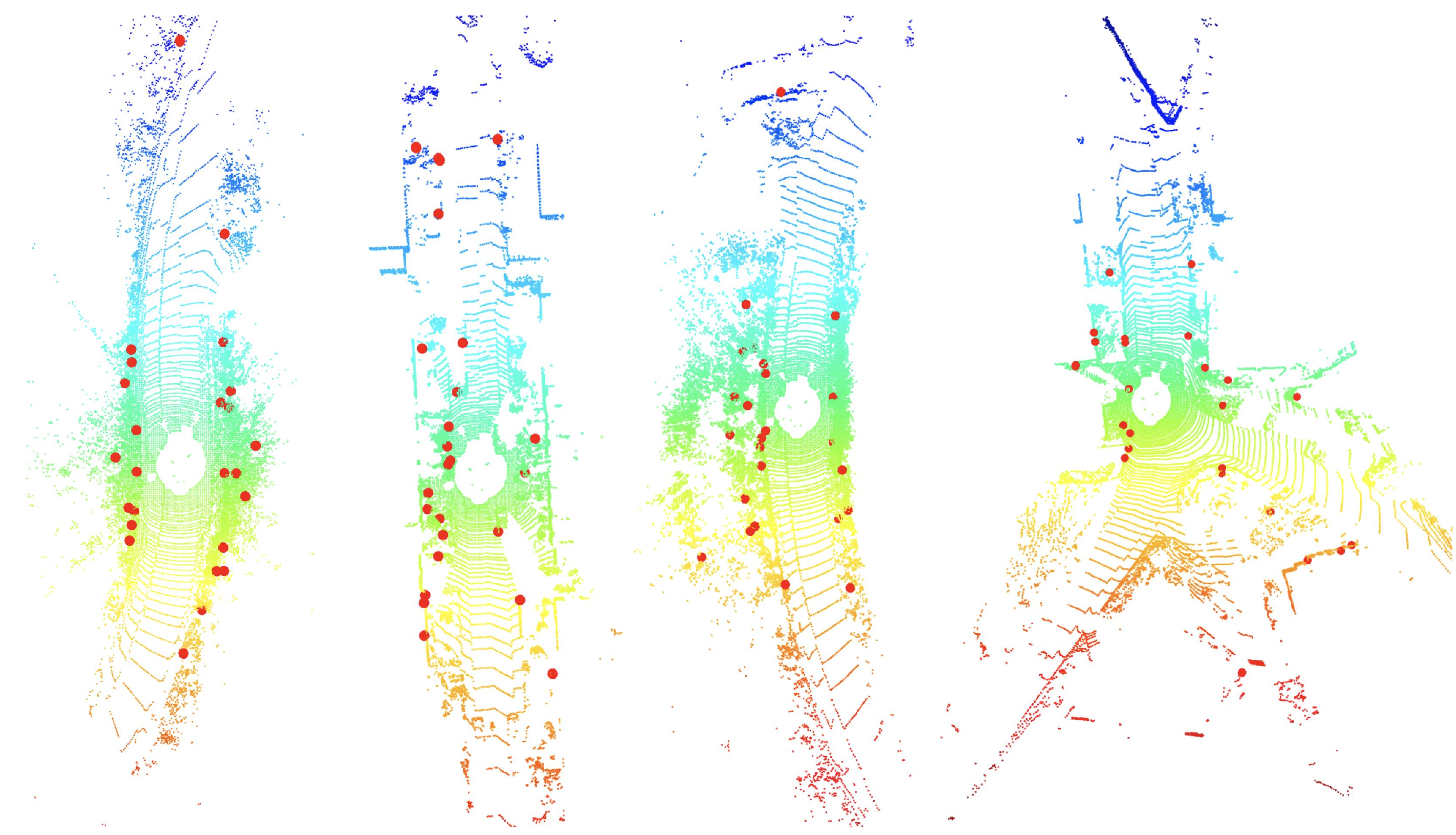}
    \\
    \includegraphics[width=18cm]{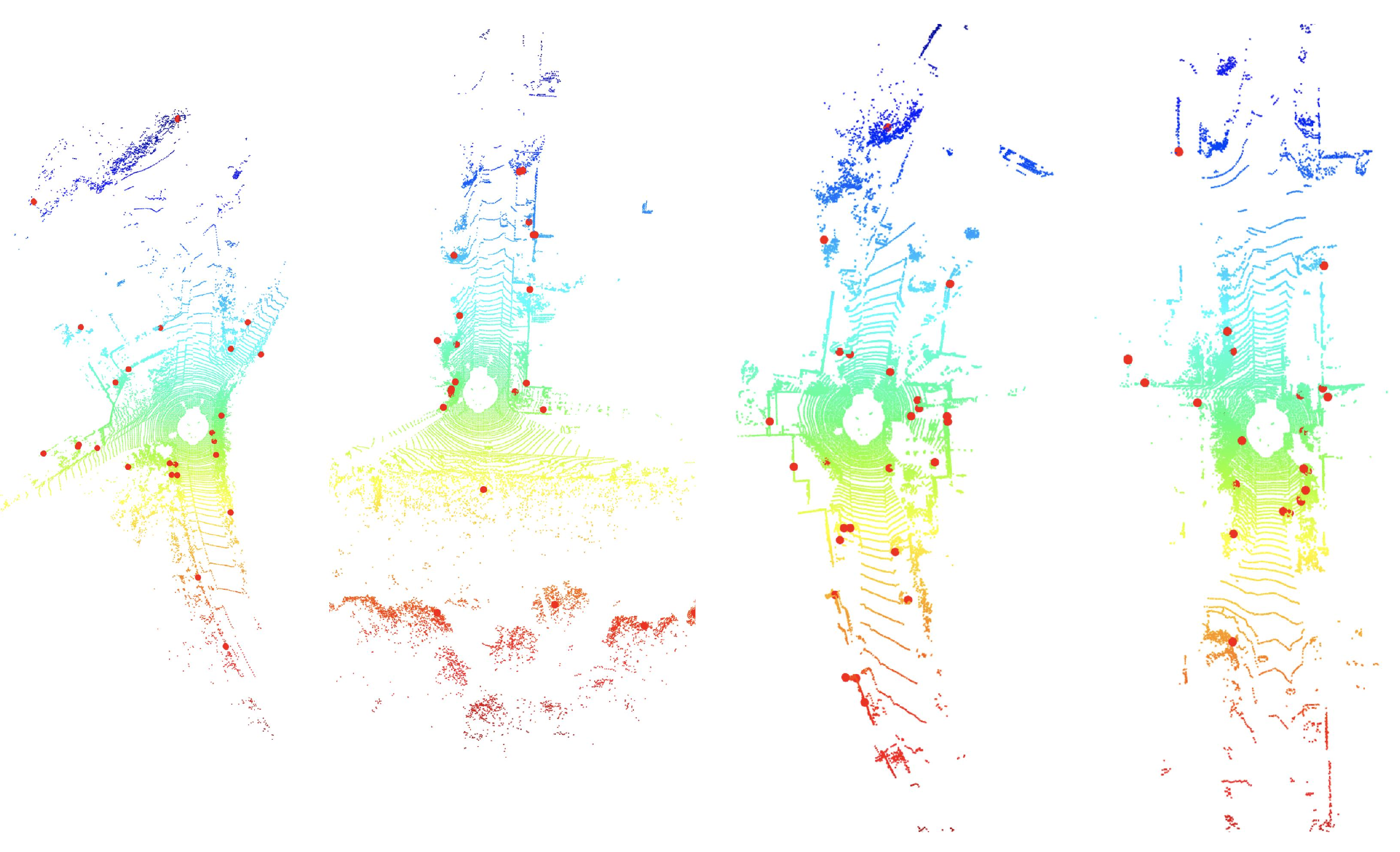}
     \caption{Visualization of keypoints on the KITTI dataset. Best view with color and zoom-in.}
    \label{fig:vis_kitti}
\end{figure*}

We show some challenging registration results in Figure~\ref{fig:vis_registration} and more visualizations of detected keypoints on 3DMatch, ETH, KITTI in Figure~\ref{fig:vis_3dmatch}, \ref{fig:vis_eth}, \ref{fig:vis_kitti}, respectively.


